\documentclass[acmsmall,authorversion,nonacm]{acmart}
\usepackage[utf8]{inputenc}
\usepackage{hyperref}
\usepackage{graphicx}
\usepackage{amsmath}
\usepackage{comment}
\usepackage{tikz}
\usetikzlibrary{positioning, calc}
\usepackage{subcaption}

\usepackage{multirow}

\AtBeginDocument{%
  \providecommand\BibTeX{{%
    Bib\TeX}}}

\AtBeginDocument{%
  \providecommand\BibTeX{{%
    \normalfont B\kern-0.5em{\scshape i\kern-0.25em b}\kern-0.8em\TeX}}}

\begin{document}

\title[Eyettention: Attention-based Human Scanpath Prediction]{Eyettention: An Attention-based Dual-Sequence Model for Predicting Human Scanpaths during Reading}
\titlenote{This is a pre-print of an article to appear in Proceedings of the ACM on Human-Computer Interaction. ETRA 2023.}

\author{Shuwen Deng}
\authornote{Both authors contributed equally to this research.}
\email{shuwen.deng@uni-potsdam.de}
\orcid{0000-0002-0185-2825}
\author{David R. Reich}
\authornotemark[2]
\orcid{0000-0002-3524-3788}
\email{david.reich@uni-potsdam.de}
\affiliation{%
  \institution{University of Potsdam}
  \city{Potsdam}
  \country{Germany}
}

\author{Paul Prasse}
\orcid{0000-0003-1842-3645}
\affiliation{%
  \institution{University of Potsdam}
  \city{Potsdam}
  \country{Germany}
}
\author{Patrick Haller}
\orcid{0000-0002-8968-7587}
\affiliation{%
  \institution{University of Zurich}
  \city{Zurich}
  \country{Switzerland}
}
\author{Tobias Scheffer}
\orcid{0000-0003-4405-7925}
\affiliation{%
  \institution{University of Potsdam}
  \city{Potsdam}
  \country{Germany}
}
\author{Lena A. J\"{a}ger}
\email{jaeger@cl.uzh.ch}
\orcid{0000-0001-9018-9713}
\affiliation{%
  \institution{University of Zurich}
  \city{Zurich}
  \country{Switzerland and}
    \institution{University of Potsdam}
  \city{Potsdam}
  \country{Germany}
}
\renewcommand{\shortauthors}{Deng and Reich, et al.}

\begin{abstract}
Eye movements during reading offer insights into both the reader's cognitive processes and the characteristics of the text that is being read. Hence, the analysis of scanpaths in reading have attracted increasing attention across fields, ranging from cognitive science over linguistics to computer science. In particular, eye-tracking-while-reading data has been argued to bear the potential to make machine-learning-based language models exhibit a more human-like linguistic behavior. However, one of the main challenges in modeling human scanpaths in reading is their dual-sequence nature: the words are ordered following the grammatical rules of the language, whereas the fixations are chronologically ordered. As humans do not strictly read from left-to-right, but rather skip or refixate words and regress to previous words, the alignment of the linguistic and the temporal sequence is non-trivial. In this paper, we develop \emph{Eyettention}, the first dual-sequence model that simultaneously processes the sequence of words and the chronological sequence of fixations. The alignment of the two sequences is achieved by a cross-sequence attention mechanism. We show that \emph{Eyettention} outperforms state-of-the-art models in predicting scanpaths. We provide an extensive within- and across-data set evaluation on different languages. An ablation study and qualitative analysis support an in-depth understanding of the model's behavior.

\end{abstract}




\maketitle

\section{Introduction}
\label{sec:intro}
 
Eye movements during reading offer insights into both the reader's cognitive processes \citep{Rayner1998} and the characteristics of the text that is being read \citep{rayner2009}. Hence, the analysis and prediction of scanpaths in reading have attracted increasing attention from different fields, ranging from  cognitive science over linguistics to computer science. On the one hand, the analysis of eye movements in reading and the development of computational cognitive models that generate scanpaths in reading have a decades-long tradition in cognitive psychology \citep{reichle2003ezreader, engbert2002dynamical} and psycholinguistics \citep{engelmann2013framework}. The goal of this line of research is  to simulate human reading behavior in order to ultimately understand the cognitive processes involved in reading and, more generally, human language processing. 
On the other hand, recent machine learning research has demonstrated that modeling scanpaths in reading bears the potential to improve a wide range of technological applications.  

In particular, eye-tracking-while-reading data has been argued to be an invaluable resource for making machine-learning-based language models exhibit a more human-like linguistic behavior, for example by regularizing neural attention mechanisms with human eye gaze \citep{barrett2018unsupervised}. First, human eye-tracking-while-reading data has been leveraged to improve the performance of neural language models on a wide range of NLP tasks such as part-of-speech-tagging~\citep{barrett2016cross}, sentiment analysis~\citep{mishra2017leveraging}, named entity recognition~\citep{hollenstein-zhang-2019-entity}, sentence compression~\citep{klerke2016}, predicting text readability~\citep{gonzalez-garduno-sogaard-2017}, generating image captions~\citep{takmaz-etal-2020-generating} and question answering \citep{sood2021multimodal}. 
Second, eye movements in reading have been used to gain insights into the inner workings of deep neural language models and how they represent linguistic knowledge. For example, a steadily increasing number of studies investigates how the syntactic representations learned by deep neural language models differ from the grammatical knowledge as it is encoded in the human mind~\citep{sood2020interpreting, hollenstein2021multilingual,hollenstein2022patterns, merkx-frank-2021-human}. 
Finally, in recent years, researchers have started to develop machine learning algorithms to infer a reader's characteristics, like their emotional state \citep{lim2020emotion}, cognitive conditions, such as ADHD \citep{deng2022detection} or dyslexia \citep{Raatikainen2021DetectionData, haller2022eye-tracking}, or their linguistic skills, such as reading comprehension capacity, or whether they are native speakers of the stimulus' language \citep{reich2022inferring, Ahn2020TowardsBehavior, Berzak2018Assessing}.

Although the different approaches differ with respect to whether they leverage eye-tracking data only for training (e.g., to adjust the model's neural attention mechanism to make its inductive bias more human-like) \citep{barrett2018unsupervised,Sood2020ImprovingAttention}, or whether eye-tracking data is also required at application time (e.g., for making inferences about how well a reader understands a given text \citep{reich2022inferring}), all of the above mentioned use cases suffer from eye-tracking data scarcity. Despite the potential, many researchers are reluctant to set up and conduct large-sample eye tracking experiments as they require considerable resources---from trained personnel needed to supervise participants on a one-to-one basis, to eye-tracking lab facilities and sometimes complicated processes of obtaining ethics approval. Researchers are well aware that  data scarcity is a major bottleneck for the field and substantial efforts are being  made to address this problem by collecting eye-tracking-while reading data in large-scale multi-lab projects \citep{cost, siegelman2022expanding}. 

In this paper we present a less resource intense way to overcome the scarcity of eye-tracking data: We develop a model that generates synthetic eye-tracking-while-reading data for any given stimulus sentence/text. Seminal work by \citep{Sood2020ImprovingAttention} who used synthetic data generated by E-Z Reader, a cognitive model of eye-movement-control in reading \citep{reichle2003ezreader}, demonstrated that synthetic eye movement data bears the potential to not only improve neural language models on standard NLP tasks such paraphrase generation, sentence compression or visual question answering~\citep{Sood2020ImprovingAttention,sood2021multimodal}, but also opens the possibility to use (synthetic) eye-tracking data as model input at application time in use cases where no data is recorded at application time. 

Although generative models of eye movements in reading have a decades-long  history in cognitive psychology, and more recently have also received attention from machine learning researchers, all existing approaches suffer from one important conceptual and technical limitation: they model eye movements in reading as a standard (i.e., single axis) sequence problem. However, one of the key properties of eye-tracking-while-reading data is its dual-sequence nature: 
The words are ordered following the grammatical rules of the language (\textit{linguistic sequence axis}), whereas the fixations on these words are chronologically ordered (\textit{temporal sequence axis}). As humans do not strictly read from left-to-right, but rather skip or re-fixate words and regress to previous words, the alignment of the linguistic and the temporal sequence axes poses a major architectural challenge. Neither  standard sequence models (i.e., with a single input sequence axis), nor encoder-decoder architectures,  that align an input with an output sequence, can handle the \textit{dual-sequence input} structure of eye movements in reading. 
Existing approaches either disregard the linguistic stimulus and only focus on the properties of the scanpath, or aggregate the scanpath by computing so-called reading measures for each word of the linguistic stimulus such as total fixation time (i.e., the sum of the durations of all fixations on a given word), or regression probability (the proportion of readers who initiate a regression from a given word). This data aggregation procedure comes at the cost of potentially relevant sequential information and limits the possible use cases for which the model can be deployed.  In this paper, we develop \emph{Eyettention}, the first dual-sequence model that  simultaneously processes the sequence of words and the chronological sequence of fixations. The alignment of the two sequences is achieved by a cross-sequence attention mechanism.

The contributions of this work are manifold:

\begin{itemize}
    \item We develop the first end-to-end trained dual-sequence encoder-encoder architecture for next fixation prediction, and improve upon the current state-of-the-art.
    \item We propose to align the linguistic axis of the stimulus sentence with the temporal axis of the fixation sequence using a local cross-attention mechanism mimicking the human visual field. 
    \item We present extensive experiments to evaluate our proposed model in a range of different application scenarios, within and across data sets, and for two typologically different languages with different scripts (alphabetic vs logographic). 
    \item We conduct an ablation study to investigate the impact of the different model components and input features. 
    \item We provide further qualitative and quantitative inspections of our model's behavior. 
\end{itemize}

The remainder of this paper is structured as follows. Section~\ref{sec:related-work} summarizes the related work, Section~\ref{sec:problem-setting} states the problem setting, Section~\ref{sec:eyettention} presents the dual-sequence Eyettention model, which is evaluated against the state-of-the-art and further inspected in Section~\ref{sec:experiments}. Section~\ref{sec:discussion} discusses the results, and Section~\ref{sec:conclusion} concludes. \looseness-1

\section{Related Work}
\label{sec:related-work}
In cognitive psychology, two computational models of eye movement control have been dominating the field for the past two decades: The E-Z reader model \citep{reichle2003ezreader} and the SWIFT model \citep{engbert2005swift}. Both models receive text as input and predict fixation location and duration, and allow for the influence of linguistic variables such as lexical frequency or predictability. However, these models were specifically designed to explain observed average eye movement phenomena at the group-level based on cognitively plausible mechanisms, but disregarding individual differences exhibited by readers. Since these methods are hardly evaluated on unseen data, their generalization ability remains unclear.

To reduce these limitations, reader-specific machine learning models were developed to predict eye movement patterns of individual readers on unseen text, the first being~\citet{nilsson2009learning}. They deployed a  logistic regression model trained on hand-engineered features extracted from eye-tracking data, such as word length, frequency, and saccade distance, to predict a reader's next fixation. 
In a follow-up study, they included additional linguistic features, e.g., surprisal and n-gram probability, to further improve the model's performance~\citep{nilsson2011entropy}. Alongside, they proposed a new method for evaluating probabilistic models by computing the entropy assigned by the model to observed eye gaze data.

Recent research draws inspiration from NLP sequence labeling tasks, and treats eye movement prediction during reading in a similar way. Specifically, the models predict the probability of each word in a sentence being fixated.
For example, \citet{hahn2016modeling,hahn2023modeling} proposed an unsupervised sequence-to-sequence architecture, with the objective of reconstructing the entire sentence using as few fixated words as possible. A labeling network was used to determine whether the next word should be fixated or not. However, the model showed limited predictive performance compared to supervised approaches. \citet{wang2019new} proposed an approach that uses a combination of convolutional neural networks (CNN), long short-term memory (LSTM) layers and a conditional random fields layer to predict the fixation probability of each word in a sentence.
However, solely returning per-word probabilities poses crucial limitations for these approaches: The model does not account 
for the chronological order in which the words are fixated, and is unable to predict  important aspects of eye movement behavior, such as regressions and refixations. 

In sum, all previous approaches either worked with aggregated representations of one or both sequence axes, or even disregarded one or the other altogether, limiting the predictive power of the respective method.

\section{Problem Setting}
\label{sec:problem-setting}
Let $\mathbf{W} = \langle w_1,\dots,w_m\rangle$ be a sentence, where for English-language texts, each word $w_j$ is an actual word whereas for Chinese texts, we use $w_j$ to represent characters. Further, let $\langle \mathbf{f}_1,\dots,\mathbf{f}_{i-1}\rangle$ be an initial part of the complete scanpath $\langle \mathbf{f}_1,\dots,\mathbf{f}_n\rangle$ of fixations. Here,  $\mathbf{f}_i=(f_i, \mathbf{F}_i)$ contains the fixation location $f_i$, which is a word index $j$, $1\leq j\leq m$, and a tuple $\mathbf{F}_i$ consisting of fixation duration and relative landing position within the word. For any  initial part of a scanpath~$\langle \mathbf{f}_1,\dots,\mathbf{f}_{i-1}\rangle$ on sentence $\mathbf{W}$, the goal is to predict the next word fixation location~$f_i$. We will estimate a likelihood function $P(f_i\vert \mathbf{W},\mathbf{f}_1,\dots,\mathbf{f}_{i-1})$ and will evaluate models in terms of the likelihood which they assign to observed scanpaths. Crucially this problem setting is reader-unspecific, that is, we aim for a model that generalizes to novel readers not seen during training, and hence can be deployed in user-independent use-cases as a one-size-fits-all model. For user specific use cases, we further investigate a variation of this problem setting where we additionally use the reader id $k$. The likelihood then changes to $P(f_i\vert k, \mathbf{W}, \mathbf{f}_1,\dots,\mathbf{f}_{i-1})$.

\section{Eyettention}
\label{sec:eyettention} 
Inspired by both human-reading behavior and the idea of encoder-decoder architectures that map an input sequence to an output sequence~\cite{sutskever2014sequence}, we propose an encoder-encoder architecture---\emph{Eyettention}---that simultaneously processes two input sequences: the initial part of the scanpath and the stimulus sentence shown during recording. They follow two unaligned sequence axes, which we will refer to as \emph{temporal sequence axis} (fixation sequence) and \emph{linguistic sequence axis} (word sequence), respectively. The representations of the two input sequences are then aligned by means of a neural attention mechanism~\cite{luong2015effective}, which we will refer to as \emph{cross-attention}. The output of the model is a probability distribution of all possible saccade ranges including progressive (i.e., positive) and regressive (i.e., negative) saccade distances measured in the number of words.

\subsection{Model architecture}
\label{sec:model}
The model processes the stimulus sentence and the initial part of the recorded fixation sequence.
It consists of two encoders, the \emph{Word-Sequence Encoder}~(see~Section~\ref{sec:sentence-encoder}) that processes a sequence of words and the \emph{Fixation-Sequence Encoder}~(see~Section~\ref{sec:fixation-sequence-encoder}) processing the sequence of fixations.
We prepend a special fixation $\mathbf{f}_0$ for \texttt{[CLS]} to the fixation sequence 
$\langle \mathbf{f}_1, \dots, \mathbf{f}_{i-1}\rangle$ to denote the beginning of a fixation sequence.

\begin{figure*}[!ht]
	\centering
	  \includegraphics[width=\textwidth,keepaspectratio,trim={0.2cm 0 0.0cm 0}, clip]{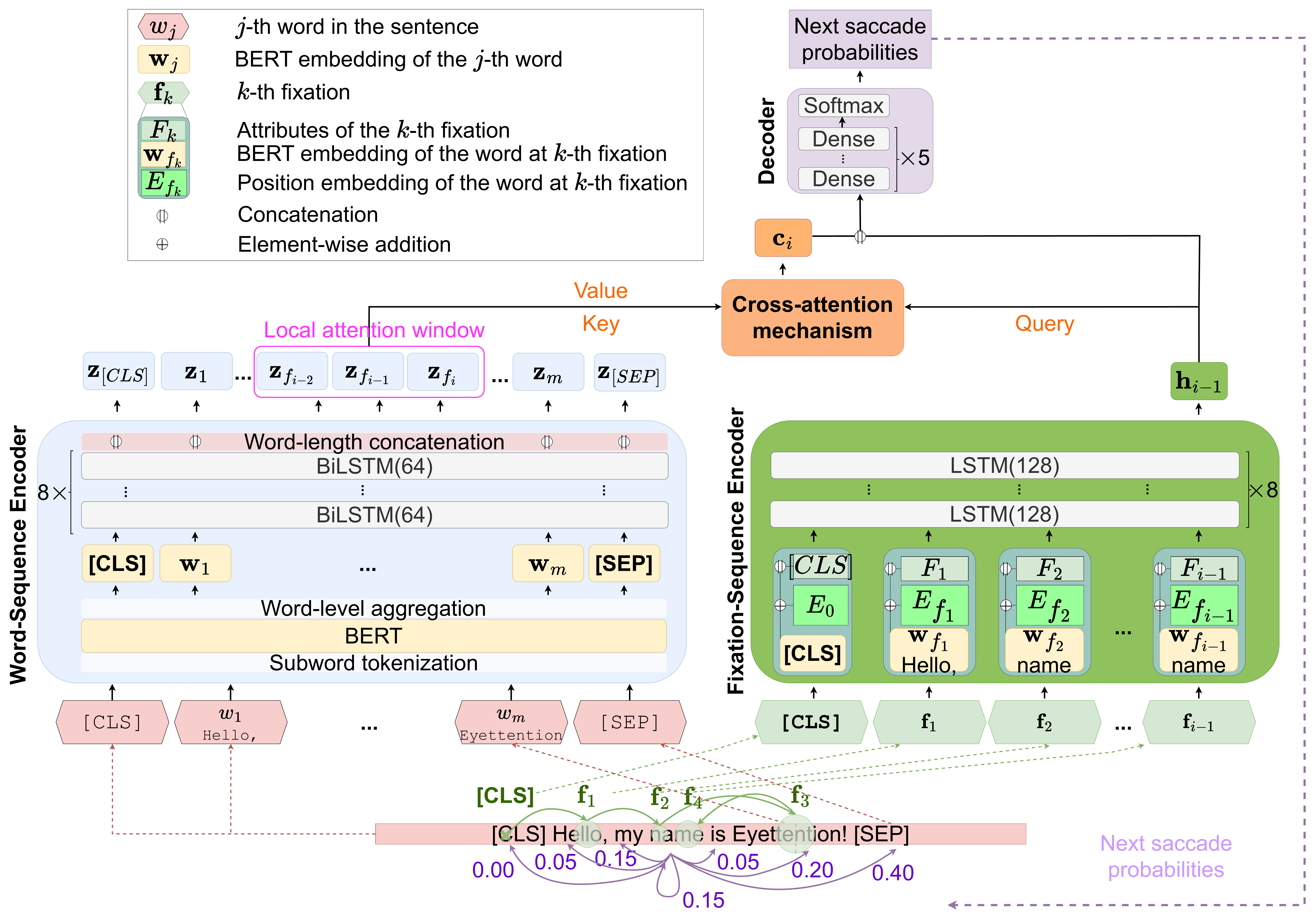}
	  \caption{\emph{Eyettention model.} Overview of our proposed Eyettention model. The model consists of three main parts: the Word-Sequence Encoder shown in the blue box on the left (described in  Section~\ref{sec:sentence-encoder}), the Fixation-Sequence Encoder shown in the green box on the right (described in Section~\ref{sec:fixation-sequence-encoder}), and the cross-attention mechanism in the red box on the top (described in Section~\ref{sec:attention}). The Decoder predicting the next fixation location is shown in the purple box and described in Section~\ref{sec:decoder}.
	  }
	\label{fig:Eyettention}
\end{figure*}

\subsubsection{Word-Sequence Encoder: Embedding the linguistic sequence}
\label{sec:sentence-encoder}
The Word-Sequence Encoder, depicted in Figure~\ref{fig:Eyettention} processes a sequence of words and extracts embeddings on a word level. 
We use a large state-of-the-art pre-trained language model, BERT~\cite{devlin2018bert}, to extract contextualized word embeddings. Since BERT uses sub-word tokens as input, we split words into sequences of sub-words. The final embedding of a word is computed by summing up the embeddings of its constituting sub-word tokens; we refer to the aggregated BERT embedding of word $w_j$ as $\mathbf{w}_j$. Our Word-Sequence Encoder consists of eight bidirectional LSTM layers (BiLSTMs)~\cite{hochreiter1997long} with $64$ units each. To avoid overfitting, a dropout layer with a dropout probability of 0.2 is applied after the BERT embedding, and a dropout layer with a probability of 0.4 after each of the first seven BiLSTM layers. The output of the BiLSTM is concatenated with the word length feature to result in the final word embedding $\mathbf{z}_j$. 

\subsubsection{Fixation-Sequence Encoder: Embedding the Temporal Sequence}
\label{sec:fixation-sequence-encoder}
The Fixation-Sequence Encoder processes a sequence of fixations together with the embeddings of the fixated words. Each fixation $\mathbf{f}_k$ consists of three attributes: word index $f_k$ of the fixated word, fixation duration and landing position within a word. We denote the last two attributes together as $\mathbf{F}_k$. The Fixation-Sequence Encoder consists of eight LSTM layers with  $128$ units each. To avoid overfitting, a dropout layer with a dropout probability of 0.4 is applied after each of the first seven LSTMs. The input at each time step is an embedding of the fixated word concatenated with both the z-score normalized fixation duration and landing position within the word. The embedding of a fixated word is calculated by the sum of the non-contextual BERT embedding (retrieved from the look-up table) of the fixated word $\mathbf{w}_{f_k}$\footnote{We use the same notation $\mathbf{w}$ for both contextual and non-contextual BERT embeddings to simplify the notation usage.} and a trainable position embedding\footnote{Note that the trainable position embedding enables the model to learn the incoming saccade length.}~\cite{gehring2017convolutional} that encodes the word index $f_k$. The output $\mathbf{h}_{i-1}$ of the Fixation-Sequence Encoder is the final hidden state of the last LSTM layer.

\subsubsection{Cross-Attention Mechanism for the Alignment of the Input Sequences}
\label{sec:attention}
The cross-attention mechanism uses the output of the two encoders described above, where the sequence for the Word-Sequence Encoder is ordered according to the \textit{linguistic sequence axis} and the sequence of the Fixation-Sequence encoder is ordered chronologically according to the \textit{temporal sequence axis}.
To align these two axes, we propose an attention-based alignment scheme, which fuses the two sequences together, allowing for cross-attention between the fixation sequence and the linguistic stimulus.

Humans typically process information within the effective visual field (perceptual span), which usually extends a few words to the left and right of a fixation location~\cite{Rayner1998}. Therefore, instead of attending to the global text input, we use a windowed Gaussian version of the dot-product attention mechanism~\cite{luong2015effective} to aggregate local information at each time step.

Let $\mathbf{h}_{i-1}$ be the output of the Fixation-Sequence Encoder at the time step $i-1$. The attention weights $a_i(n)$ for the $n$-th word-embedding, for a window of size $D$, are computed based on the query $\mathbf{h}_{i-1}$ and the Word-Sequence Encoder output $\mathbf{z}$~(served as keys), as follows:
\begin{equation*}
\displaystyle
\mathbf{a}_{i}(n) = \frac{\exp\ (\mathbf{h}_{i-1}^T W_{a}\mathbf{z}_{n})}{\sum_{k=f_{i-1}-D}^{f_{i-1}+D}\exp\ (\mathbf{h}_{i-1}^T W_{a}\mathbf{z}_{k})} \underbrace{\exp(-\frac{(n-f_{i-1})^{2}}{2\sigma^{2}})}_\text{Gaussian kernel}\ ,
\end{equation*}
where $W_{a}$ represents the learnable weights to align $\mathbf{h}_{i-1}$ and $\mathbf{z}_k$.  The standard deviation of the Gaussian smoothing is set to $\sigma=D/2$. The final output of the cross-attention is the vector $$\mathbf{c}_{i}=\sum_{k=f_{i-1}-D}^{f_{i-1}+D}\mathbf{a}_{i}(k)\mathbf{z}_k.$$

Conceptually, this means that the neural representation of the scanpath up to the current time step is used to weigh the contextualized representations of the words in the stimulus sentence according to their relevance for deciding where to fixate next. 
We `help' the model to attend to relevant words by constraining its `perceptual span' the local attention window.

\subsubsection{Decoder: Next Fixation Prediction}
\label{sec:decoder}
The output of the cross-attention $\mathbf{c}_{i}$ is concatenated with the output of the Fixation-Sequence Encoder $\mathbf{h}_{i-1}$. It is then fed into a four-layer fully-connected network with rectified linear unit activation functions. To avoid overfitting, a dropout layer with a probability of 0.2 is applied before each of the fully-connected layers. Finally, to calculate a probability distribution over all possible saccade ranges, a softmax layer is applied. We define the set of possible saccade ranges based on the maximal sentence length $M$ observed in the training data as $\{-M+1, -M+2, \ldots, M\}$,
 where positive and negative saccade ranges represent progressive and regressive saccades, respectively. An additional class is added to denote the end-of-scanpath. Hence, we can treat our problem as a multi-class classification task. Note that predicting the next saccade range is equivalent to predicting the next fixation location directly, since we can calculate the next fixation location as the sum of the current fixation location and the saccade range. We choose to model the probability distribution of the next fixation location via saccade ranges, as this allows us to predict fixation locations independently of the sentence length. The model generates a new scanpath by iterative sampling from the distribution of the next fixation location $P(f_i\vert \mathbf{W},\mathbf{f}_0,\dots,\mathbf{f}_{i-1})$. The procedure stops when the end-of-scanpath or a predefined maximum scanpath length has been reached.

\subsection{Reader-Specific Eyettention$_\text{reader}$}
\label{sec:reader-specific-eyettention}
One of the application scenarios that we will consider in our experiments (see Section~\ref{sec:new-sentence}), investigates the model's ability to generate reader-specific scanpaths. This application scenario allows us to introduce a trainable reader identifier embedding with an embedding size of $d_{emb}$ to our model. We refer to this version of our model as Eyettention$_\text{reader}$. The embedding is concatenated to the fixation representation $\mathbf{f}_k$ in the Fixation-Sequence Encoder.

\subsection{Model Configuration and Optimization} 
\label{sec:model-config-and-optimization}
 
Given a training data set $\mathcal{D}=\{(\mathbf{S^1}, \mathbf{W^1}), (\mathbf{S^2}, \mathbf{W^1}), \ldots,(\mathbf{S^j}, \mathbf{W^2}), (\mathbf{S^{j+1}}, \mathbf{W^2}),\ldots (\mathbf{S^E}, \mathbf{W^D})\}$, where $\mathbf{S^d}$
indicates the scanpaths recorded during reading sentence $\mathbf{W^d}$, our model is trained by minimizing the average of negative log-likelihood in an end-to-end fashion.  Note that each sentence can be associated with scanpaths of more than one reader. The log-likelihood of a scanpath can be reformulated as the sum over the log-likelihoods for each fixation. 

$$L = -\frac{1}{\lvert \mathcal{D} \rvert}\sum_{d=1}^{\lvert \mathcal{D} \rvert}\log p(\mathbf{S^d}\vert\mathbf{W^d}) =  -\frac{1}{\lvert \mathcal{D} \rvert} \sum_{d=1}^{\lvert \mathcal{D} \rvert} \left(\frac{1}{\lvert \mathbf{S^d} \rvert}  \sum_{i=1}^{\lvert \mathbf{S^d} \rvert}\log p(f_i\vert\mathbf{W^d},\mathbf{f}_0,\dots,\mathbf{f}_{i-1})\right).$$

The network parameters are optimized using the Adam optimizer~\cite{kingma2014adam} with a learning rate of 1e-3. We train the models for 1000 epochs using early stopping (patience of 20 on a validation set sampled from the training data), with a batch size of 256. All neural networks are trained using the PyTorch~\cite{pytorch2019paszke} library on an NVIDIA A100-SXM4-40GB GPU using the NVIDIA CUDA platform. The code we used to train and evaluate our models is available online
\footnote{https://github.com/aeye-lab/Eyettention}.

\section{Experiments}
\label{sec:experiments}
To evaluate the performance of our model against cognitive and machine learning-based state-of-the-art methods of scanpath generation, we investigate the problem setting described in Section~\ref{sec:problem-setting} using a range of publicly available eye-tracking-while-reading data sets. 
We distinguish between within- and across-data set evaluation. For the within-data set evaluation we consider three different train/test splits that represent different use case scenarios: We evaluate how the models perform when predicting scanpaths of (i) known readers on novel sentences (\textit{New Sentence} split), (ii)  novel readers on known sentences (\textit{New Reader} split), and (iii) novel readers on novel sentences (\textit{New Reader / New Sentence} split).  
In the cross-data set evaluation, we assess the models' ability to generalize to a novel data set that differs with respect to the experimental set-up (e.g. eye tracking hardware, sampling rate) as well as the properties of the stimuli.

In the remainder of this section, we describe the data sets, the reference methods, the hyperparameter tuning, and the evaluation protocols and results for the different evaluation protocols.

\subsection{Data Sets}
\label{sec:datasets}
We use four corpora to train, tune and/or evaluate our proposed model and the reference methods (see Section~\ref{sec:ref-methods}). The different data sets differ with respect to the stimulus language and script (English vs Chinese), as well as stimulus layout and eye tracking hardware. Descriptive statistics for each of the data sets are presented in Table~\ref{tab:descriptive-corpora-stats}.

The \emph{Beijing Sentence Corpus} (BSC) is a Chinese sentence corpus recorded by 
~\citet{pan2021bsc}. All readers are native speakers of Chinese.
The \emph{Corpus of Eye Movements in L1 and L2 English Reading}~(CELER) presented by 
~\citet{berzak2022celer} is an English sentence corpus. CELER includes data from native (L1) and non-native (L2) speakers of English. For our study, we only include the L1 data (CELER~L1). 
Hollenstein~et al. collected two English sentence corpora, the \emph{Zurich Cognitive Language Processing Corpus} (ZuCo)~\cite{hollenstein2018zuco} and the \emph{Zurich Cognitive Language Processing Corpus 2.0} (ZuCo~2.0)~\cite{hollenstein2019zuco2}. All readers were native English speakers and included in two distinct reading paradigms: ``task-specific'' and ``natural reading''. In our study, we only include the ``natural reading'' subset (ZuCo/ZuCo~2.0~NR).

\begin{table}[h!]
  \small
  \caption{Descriptive statistics of the four eye-tracking-while-reading corpora used for model training and evaluation: BSC~\cite{pan2021bsc}, CELER~\cite{berzak2022celer}, ZuCo~\cite{hollenstein2018zuco}, and ZuCo~2.0~\cite{hollenstein2019zuco2}. The number of words per sentence is reported using the mean $\pm$ standard deviation.}
    \centering
    \begin{tabular}{l|l|c|c|c|l}
    \toprule
           & &\# Unique & \# Words per &  &  \\
         Dataset & Eye-tracker & sentences & sentence & \# Readers  & Language \\ \hline
         BSC~\cite{pan2021bsc} & Eyelink~II (500~Hz)& $150$ &$11.2\pm1.6$ & $60$ & Chinese  \\
         CELER~L1~\cite{berzak2022celer} & Eyelink~1000 (1000~Hz) & 5456 & 11.2 $\pm$ 3.6 & $69$ & English \\
         ZuCo~NR~\cite{hollenstein2018zuco} & EyeLink~1000~Plus (500~Hz)& $700$& 19.6 $\pm$ 9.8& $12$&English\\
         ZuCo~2.0~NR~\cite{hollenstein2019zuco2} &EyeLink~1000~Plus (500~Hz) & $349$&$19.6\pm8.8$& $18$&English\\
         \bottomrule
    \end{tabular}

    \label{tab:descriptive-corpora-stats}
\end{table}

\subsection{Reference Methods}
\label{sec:ref-methods}
In order to benchmark our model against the state-of-the-art approaches from the different research traditions
, we compare our model's performance with the two most widely used cognitive models of eye movements in reading, namely the E-Z~reader~model~\cite{reichle2003ezreader,rayner2007chineseezreader} and the SWIFT model~\cite{rabe2021bayes}, as well as two state-of-the-art machine-learning-based models, introduced in Section~\ref{sec:related-work} (\citet{nilsson2009learning, nilsson2011entropy}.
To put the performance scores of the different models into perspective, we also include two trivial  baseline models: The \textit{Uniform model} predicts fixations to be independently and uniformly distributed over the sentence; the \textit{Train-label-dist} model samples the saccade range from the training label distribution. We further compare to human-level inter-reader scanpath similarity, which we will call \textit{Human}.
\subsection{Evaluation Metrics}
To assess model performance, we use negative log-likelihood (NLL), a widely accepted probability-based metric commonly employed to evaluate generative models for image synthesis~\citep{theis2016a}. NLL measures the average of the negative log-probability scores assigned by the model to each individual fixation in the scanpath, which reflects how confidently the model predicts each fixation location. To account for different scanpath lengths, the NLL is normalized by the length of the scanpath. This normalization accounts for the fact that longer scanpaths contain more fixations and are thus more challenging to predict accurately. A lower NLL indicates better predictive performance. NLL allows us to evaluate a model's ability to capture the variability in human gaze behavior, whereas sampling-based metrics such as accuracy only assess exact fixation points~\cite{nilsson2010towards, kummerer2021state}. Moreover, according to \citep{kummerer2021state}, the process of evaluating each fixation conditioned on the preceding scanpath closely aligns with the underlying biological processes of scanpath generation.
For both \textit{Human} and E-Z reader, no likelihood of the next fixation location is available, hence we cannot compare them in terms of NLL.

Besides NLL, we calculate the similarity between human and model-generated scanpaths. For this we will employ the normalized Levenshtein distance (NLD). The Levenshtein distance (LD)~\cite{levenshtein1966binary} is the minimal number of additions, deletions and substitutions needed to transform a human word-index sequence $S$ into the generated word-index sequence $T$. We normalize the LD by the maximum length of the two sequences, hence NLD is defined as $NLD(S,T) = LD(S, T)/\max(\lvert S \rvert, \lvert T\rvert)$. For each human scanpath, we generate a new scanpath from the model based on the same sentence. 
To compare against a \textit{Human} baseline in terms of NLD, we calculate the NLD between 
any human scanpath, and a randomly sampled one, recorded reading the same sentence. 
Two other commonly used metrics for measuring scanpath similarity are MultiMatch~\cite{jarodzka2010vector} and ScanMatch~\cite{cristino2010scanmatch}. However, the validity of these metrics has been questioned in a recent study~\cite{kummerer2021state}. For both metrics, it has been shown that incorrect models can systematically score higher than the ground truth model, but the issue is more severe for ScanMatch than for MultiMatch in their experimental results. Therefore, we only present the results of the MultiMatch metric in Appendix~\ref{sec:multimatch}.
To generate scanpaths with our model, both fixational attributes, i.e. landing position and fixation duration, are set to 0 as they are not available during the generation process.  Statistical tests were performed using a two-tailed t-test, with $p < 0.05$.
\subsection{Hyperparameter Tuning}
\label{sec:hp-tuning}
To find the best model configuration (number of LSTM/BiLSTM layers, number of dense layers, etc.), we  are performing a hyperparameter tuning by training on the CELER~L1 data set and validating the model on the held-out data set ZuCo~2.0~NR. The held-out tuning set is disjoint from the training data and differs with respect to the recording setup and the characteristics of presented stimuli. The tuning set is discarded after the hyperparameter optimization and not used in any further experiments. 
Table~\ref{tab: hp} of the Appendix shows the used parameter grid and indicates the best found configuration which is depicted in Figure~\ref{fig:Eyettention}. In our parameter grid, we restrict the number of units for the BiLSTM to be half the number of units for the LSTM.

\subsection{Within-Data Set Evaluation}
\label{sec:within-data-eval}
\subsubsection{New Sentence Split}
\label{sec:new-sentence}
In the \textit{new sentence} split we investigate the ability of Eyettention, its reader-specific version Eyettention$_\text{reader}$ and the reference methods to generalize to novel sentences read by known readers. We thus perform 5-fold hold-out cross-validation with the train/test split being performed along the sentence ids. In each fold, we use 80\% of the sentences for training and the remaining disjoint 20\% for testing, ensuring that all models are tested on sentences not seen during training. We apply identical data splits to all models to ensure fairness in comparison. 

\textit{Results.}
\label{sec:results-new-sentence}
An overview of the results for Eyettention and the reference methods is provided in Table~\ref{tab: res_sentence}. For negative log-likelihood (NLL), our model significantly outperforms the current state-of-the-art of
~\citet{nilsson2011entropy} by roughly 12\% on the BSC and 4\% on the CELER~L1 data set, respectively. Moreover, for each of the similarity metrics, we observe significant improvements over all reference methods for both languages. While our model significantly improves the state-of-the-art across all investigated metrics, it does not quite achieve the degree of similarity observed between scanpaths generated by different human readers on the same sentence.

\textit{Results Eyettention$_\text{reader}$.} The results for Eyettention$_\text{reader}$, see Section~\ref{sec:reader-specific-eyettention}, are shown in Table~\ref{tab: subID}. We observe that individualizing the scanpath generation to specific readers significantly improves the model's performance.  Including a reader embedding of any of the investigated sizes significantly improves the model's performance compared to the bare Eyettention model. The differences between models with different reader embedding sizes were not significant.

\begin{table}[]
\caption{\emph{New Sentence split}. Models are evaluated on novel sentences using 5-fold cross-validation. The dagger $\dagger$ indicates that a model is significantly worse than the best model. The value in brackets indicates the improvement over the uniform baseline in \%. \looseness=-1}

\label{tab: res_sentence}
\begin{center}
    \begin{tabular}{l|l|l|l}
    \toprule
    Data set & Model       & NLL $\downarrow$&NLD $\downarrow$         \\\hline
    BSC   &Uniform               &5.672$\dagger$ & 0.761 $\pm$ 0.001$\dagger$\\
          (Chinese)&Train-label-dist      &2.551 $\pm$ 0.007$\dagger$ (55.03 \%)   &0.662 $\pm$ 0.002$\dagger$\\
          & E-Z Reader &--& 0.680 $\pm$ 0.000$\dagger$\\
          &\citet{nilsson2009learning}   & 2.227 $\pm$ 0.008$\dagger$ (60.75 \%)   &0.589 $\pm$ 0.003$\dagger$    \\
          &\citet{nilsson2011entropy}    & 2.110 $\pm$ 0.008$\dagger$  (62.80 \%)  &0.564 $\pm$ 0.002$\dagger$\\
          &Eyettention (Ours)        &\textbf{1.856 $\pm$ 0.007} (\textbf{67.28 \%})   & \textbf {0.545$\pm$ 0.004}\\ \cline{2-4}
          &\textit{Human} &--&\textit{0.530 $\pm$ 0.002}\\\hline
    
    CELER~L1 &Uniform               & 5.492$\dagger$ &0.781 $\pm$ 0.001$\dagger$\\
          (English)&Train-label-dist      & 2.957 $\pm$ 0.005$\dagger$ (46.16 \%) &0.673 $\pm$ 0.002$\dagger$\\
          & E-Z Reader &--&  0.660 $\pm$ 0.000$\dagger$\\
          &SWIFT       &2.725 $\pm$ 0.004$\dagger$ (49.62 \%) &0.658 $\pm$ 0.003$\dagger$\\
          &\citet{nilsson2009learning}   & 2.582 $\pm$ 0.007$\dagger$ (52.98 \%)  &0.619 $\pm$ 0.002$\dagger$    \\
          &\citet{nilsson2011entropy}    &2.380 $\pm$ 0.006$\dagger$ (56.67 \%)  & 0.596 $\pm$ 0.002$\dagger$\\
          &Eyettention (Ours)        &\textbf{2.277 $\pm$ 0.005} (\textbf{58.54 \%})   & \textbf{0.572 $\pm$ 0.002}\\ \cline{2-4}
          &\textit{Human} &--&\textit{0.536 $\pm$ 0.006}\\
    
    \bottomrule
    \end{tabular}
\end{center}

\end{table}

\begin{table*}[]
\caption{\emph{New Sentence split for the reader-specific  Eyettention$_\text{reader}$ model}. NLL of observed scanpaths on sentences given the reader ID. The dagger $\dagger$ indicates models significantly better than the models without reader embeddings. The percentages in brackets indicate the percentage improvement over the model without reader embedding. $d_{emb}$ denotes the size of the reader-specific embedding.}

\label{tab: subID}
\small
\begin{center}
    \begin{tabular}{l|l|l|l|l}
    \toprule
     & Eyettention  & \multicolumn{3}{c}{Eyettention$_\text{reader}$}  \\
      Data set &  &$d_{emb}=16$  & $d_{emb}=32$  & $d_{emb}=64$  \\\hline
    BSC   &1.856 $\pm$ 0.007 &  \textbf{1.779 $\pm$ 0.007}$\dagger$ (\textbf{4.2 \%})          &1.783 $\pm$ 0.007$\dagger$ (4 \%)  & 1.789 $\pm$ 0.007$\dagger$ (3.6 \%) \\
    
    CELER~L1   & 2.277 $\pm$ 0.005   & 2.186 $\pm$ 0.006$\dagger$ (4 \%)       & \textbf{2.183 $\pm$ 0.006}$\dagger$ (\textbf{4.1 \%})&2.186 $\pm$ 0.006$\dagger$ (4 \%) \\
        
    \bottomrule
    \end{tabular}
\end{center}

\end{table*}

\subsubsection{New Reader Split}
\label{sec:new-reader}
In the \textit{New Reader} split, we investigate the models' ability to generalize to novel readers. We perform 5-fold cross-validation with the train/test split being performed along the reader IDs such that all models are tested on novel readers not seen during training.

\emph{Results.}
An overview of the results is provided in Table~\ref{tab: res_reader}. Eyettention significantly outperforms all reference methods across all investigated metrics for both the English and the Chinese data. For NLL, our model significantly outperforms the current state-of-the-art of~\citet{nilsson2011entropy} by roughly 11\% on the BSC and 4\% on the CELER~L1 data set, respectively.
Moreover, Eyettention almost achieves human-level similarity for the BSC  data set.

\begin{table*}[]
\caption{\emph{New Reader split}. Models are evaluated on novel readers using five-fold cross-validation. The dagger $\dagger$ indicates models significantly worse than the best model. The value in brackets indicates the improvement over the uniform baseline in \%.}

\label{tab: res_reader}
\begin{center}
    \begin{tabular}{l|l|l|l}
    \toprule
    Data set & Model       & NLL $\downarrow$   & NLD $\downarrow$         \\\hline
    BSC   &Uniform               & 5.672$\dagger$ &0.761 $\pm$ 0.001$\dagger$
\\
          (Chinese)&Train-label-dist      & 2.554 $\pm$ 0.007$\dagger$ (55.03 \%) &0.662 $\pm$ 0.002$\dagger$\\
          & E-Z Reader &--& 0.677 $\pm$ 0.000$\dagger$\\
          &\citet{nilsson2009learning}   & 2.230 $\pm$ 0.008$\dagger$ (60.69 \%)  &0.589 $\pm$ 0.003$\dagger$   \\
          &\citet{nilsson2011entropy}   & 2.104 $\pm$ 0.008$\dagger$ (62.9 \%) &0.563 $\pm$ 0.002$\dagger$\\
          &Eyettention (Ours)        &\textbf{1.875 $\pm$ 0.007} (\textbf{66.95 \%})   & \textbf{0.539$\pm$ 0.002}\\ \cline{2-4}
          &\textit{Human}&--&\textit{0.530 $\pm$ 0.002}\\\hline
    
    CELER~L1 &Uniform               & 5.492$\dagger$ &0.780 $\pm$ 0.001$\dagger$\\
          (English)&Train-label-dist      & 2.961 $\pm$  0.005$\dagger$ (46.08 \%)  &0.670 $\pm$ 0.002$\dagger$\\
          & E-Z Reader &--&  0.658 $\pm$ 0.001$\dagger$\\
          &SWIFT      &2.724 $\pm$ 0.008$\dagger$ (49.60 \%)  &0.644 $\pm$ 0.002$\dagger$\\
          &\citet{nilsson2009learning}   & 2.587 $\pm$  0.007$\dagger$ (52.89 \%)  &0.617 $\pm$ 0.002$\dagger$    \\
          &\citet{nilsson2011entropy}   & 2.373 $\pm$  0.006$\dagger$ (56.79 \%) &0.598 $\pm$ 0.002$\dagger$\\
          &Eyettention (Ours)        &\textbf{2.267 $\pm$ 0.005} (\textbf{58.72 \%})   & \textbf {0.573 $\pm$ 0.002}\\ \cline{2-4}
          &\textit{Human} &--&\textit{0.536 $\pm$ 0.006}\\
    
    \bottomrule
    \end{tabular}
\end{center}

\end{table*}

\subsubsection{New Reader / New Sentence Split.}
\label{sec:new-reader-new-sentence}
This split investigates the model's ability to generalize to novel readers and novel sentences. We conduct 5-fold random re-sampling by splitting the training and testing data along both, reader IDs and sentence IDs such that all models are tested on scanpaths generated by unknown readers on unknown sentences.

\emph{Results.}
An overview of the results is presented in Table~\ref{tab: res_sentence_reader}. Eyettention significantly improves the state-of-the-art of~\citet{nilsson2011entropy} in terms of NLL by  12~\% on the BSC and 4\% on the CELER~L1 data set, respectively. This is also reflected in the similarity measures. For CELER~L1, regarding NLL, our model is significantly better than all models in comparison. Moreover, in all similarity metrics, our model is also significantly better than all investigated models. Similar to previous results, we see that our model does not reach human similarity measures. 
\begin{table*}[]
\caption{\emph{New Reader / New Sentence}. Models are evaluated on new sentences and new readers, with 20\% of sentences and readers used for testing, which is randomly resampled 5 times. The dagger $\dagger$ shows models significantly worse than the best model. The value in brackets indicates the improvement over the uniform baseline in \%.}

\label{tab: res_sentence_reader}
\begin{center}
    \begin{tabular}{l|l|l|l}
    \toprule
    Data set & Model       & NLL $\downarrow$   &NLD$\downarrow$          \\\hline
    BSC   &Uniform               &5.672$\dagger$ &0.772 $\pm$ 0.003$\dagger$\\
         (Chinese) &Train-label-dist      &2.505 $\pm$ 0.015$\dagger$ (55.83 \%)   &0.667 $\pm$ 0.004$\dagger$\\
          & E-Z Reader &--& 0.625 $\pm$ 0.000$\dagger$\\
          &\citet{nilsson2009learning}  &2.195 $\pm$ 0.017$\dagger$ (61.31 \%)   & 0.596 $\pm$ 0.004$\dagger$  \\
          &\citet{nilsson2011entropy}    &2.093 $\pm$ 0.017$\dagger$ (63.11 \%)  &0.564 $\pm$ 0.004$\dagger$\\
          &Eyettention (Ours)        &\textbf{1.84 $\pm$ 0.017 } (\textbf{67.57 \%})   & \textbf {0.545 $\pm$ 0.004}\\ \cline{2-4}
          &\textit{Human} &--&\textit{0.524 $\pm$ 0.000}\\\hline
    
    CELER~L1 &Uniform               & 5.492$\dagger$ &0.781 $\pm$ 0.003$\dagger$\\
          (English) &Train-label-dist      & 2.949 $\pm$ 0.011$\dagger$ (46.31 \% )  &0.665 $\pm$ 0.004$\dagger$\\
          & E-Z Reader &--&  0.649 $\pm$ 0.001$\dagger$\\
          & SWIFT     &2.748 $\pm$ 0.009$\dagger$ (50.03 \%)  &0.627 $\pm$ 0.004$\dagger$\\
          &\citet{nilsson2009learning}   & 2.594 $\pm$ 0.017$\dagger$ (52.77 \%)  &0.619 $\pm$ 0.004$\dagger$    \\
          &\citet{nilsson2011entropy}    & 2.381 $\pm$ 0.013$\dagger$ (56.64 \%) &0.595 $\pm$ 0.004$\dagger$\\
          &Eyettention (Ours)        &\textbf{2.297 $\pm$ 0.011 } (\textbf{58.18 \%})   & \textbf {0.568 $\pm$ 0.004}\\ \cline{2-4}
          &\textit{Human} &--&\textit{0.542 $\pm$ 0.001}\\
    
    \bottomrule
    \end{tabular}
\end{center}

\end{table*}

\subsection{Cross-Data Set Evaluation}
\label{sec:cross-dataset-eval}
The New Reader / New Sentence results from the previous Section~\ref{sec:new-reader-new-sentence} are supposed to simulate new readers and new sentences, but presentation style and hardware setup are the same. Moreover, sentences as well as readers are from a similar corpus and background. Hence, we investigate presentation style and hardware agnostic generalizability of our proposed architecture. For this, henceforth called \emph{Cross-Data Set Evaluation}, we train on the CELER~L1~\cite{berzak2022celer} data set and evaluate our model on ZuCo~NR~\cite{hollenstein2018zuco}. We further investigate whether a pretraining/fine-tuning approach improves the performance on, as is the case with eye-tracking-while-reading applications, small data sets.

\begin{figure*}[!ht]

    \subfloat[Training on CELER, testing on ZUCO.\label{subfig:train_celer_test_zuco}]
	{%
      \includegraphics[width=0.48\textwidth,keepaspectratio]{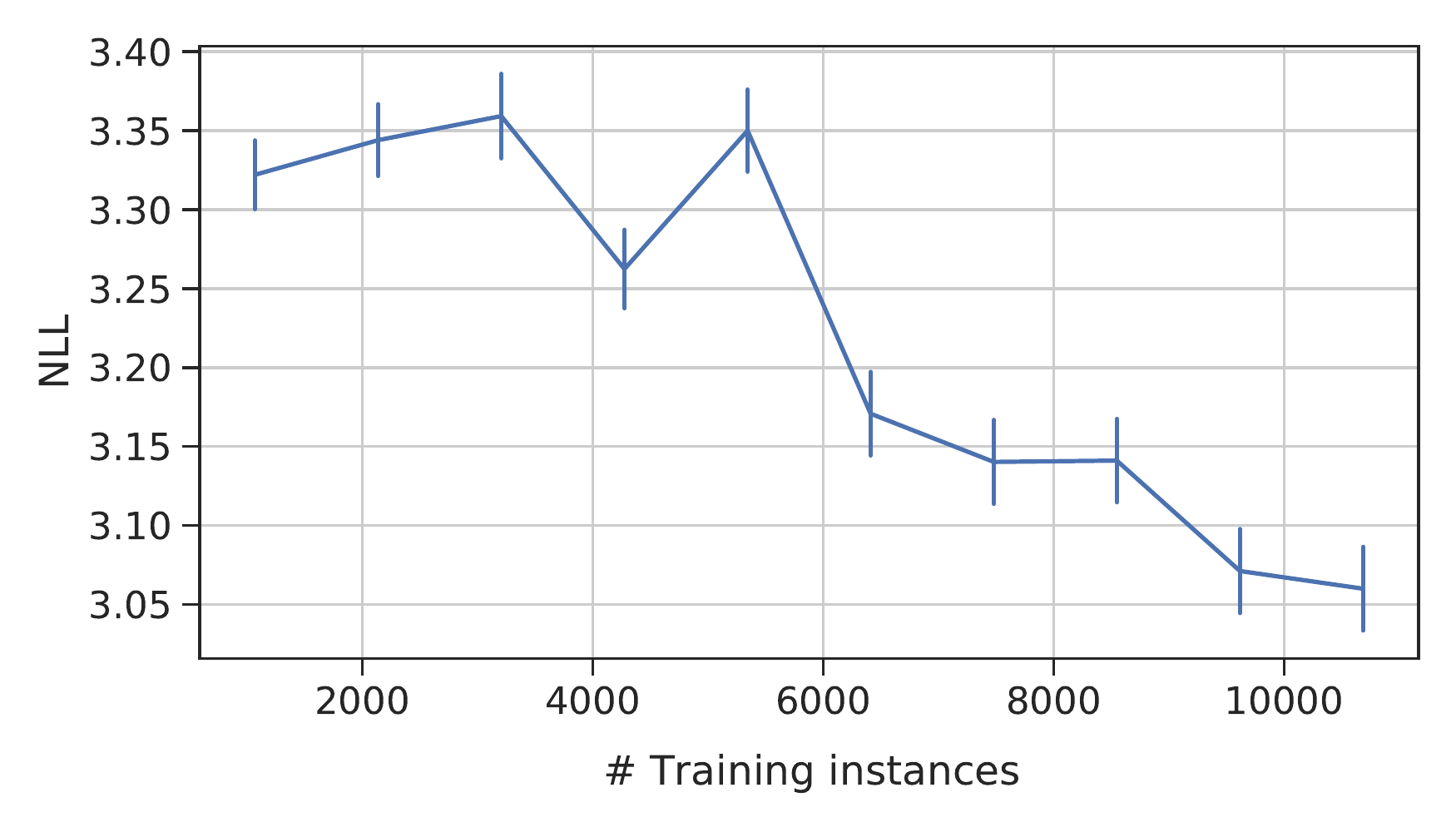}
    }
    \hfill
    \subfloat[Training/fine-tuning on ZUCO, testing on ZUCO.\label{subfig:train_celer_test_zuco_finetune}]
    {%
      \includegraphics[width=0.48\textwidth,keepaspectratio]{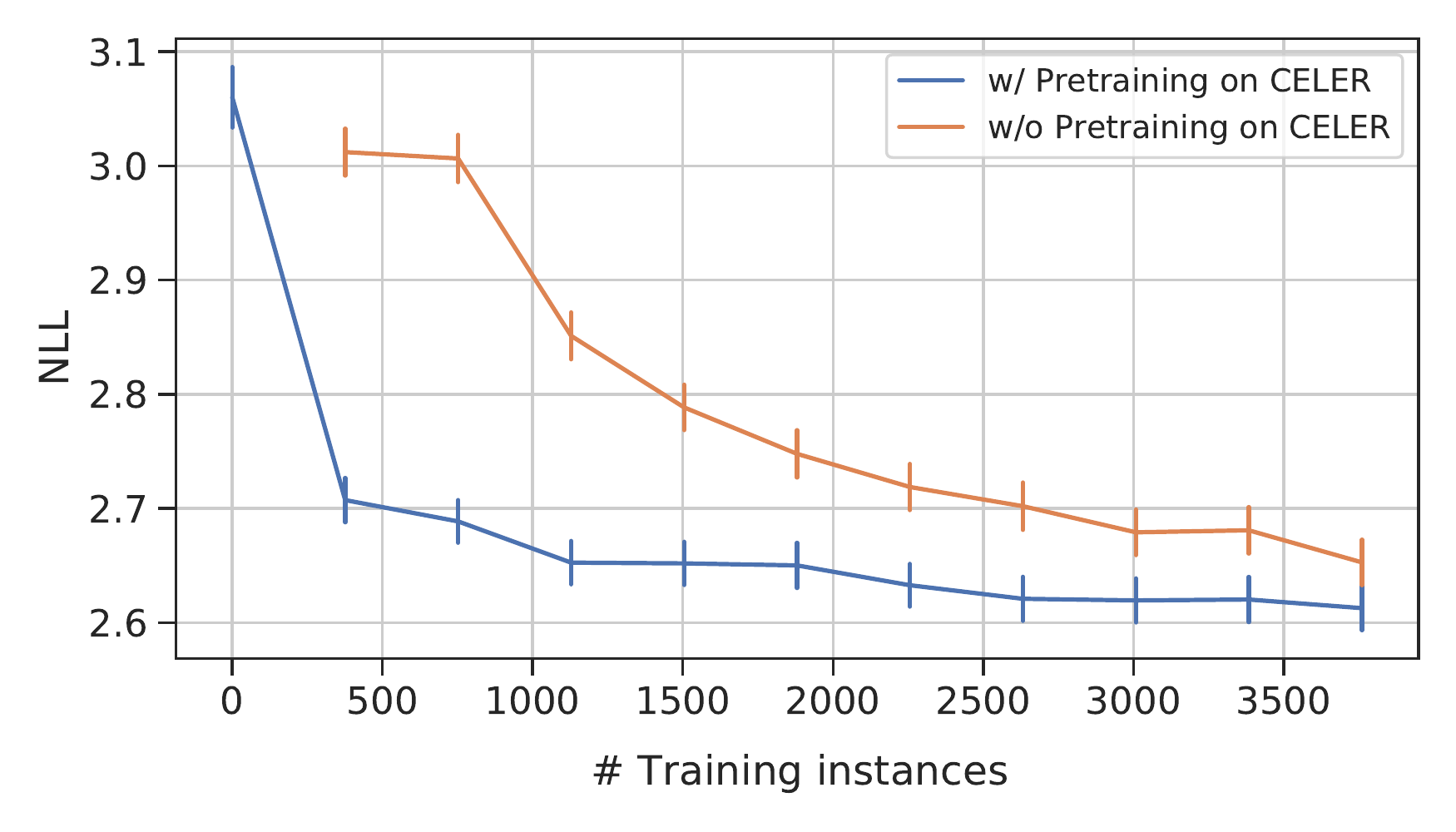}
    }

	  \caption{Impact of the proportion of the data for training and fine-tuning.}
	\label{fig:finetune}
 
\end{figure*}

\emph{Results.}
An overview of the results can be seen in Table~\ref{tab:cross-dataset-eval}. We see that our model is able to generalize reasonably well to a new data set. Next, we investigate the effect of using a small part of the test data for fine-tuning. We can see that fine-tuning Eyettention significantly improves the model's performance. In Figure~\ref{fig:finetune}, we investigate the trade-off between the number of fine-tuning instances and Eyettention's performance. We can observe that using only a few instances of the data significantly increases the model's performance compared to the model without fine-tuning.   

While the model's performance improves with more fine-tune instances, we also see that only a few hundred instances provide the majority of improvement. Overall, we observe that the pretraining approach improves Eyettention's performance compared to Eyettention trained from scratch with only ZuCo data.

\begin{table*}[]
\caption{\emph{Cross-Data Set Evaluation.} Exploring the possibility to generalize to completely different data sets.}

\label{tab:cross-dataset-eval}
\begin{center}

    \begin{tabular}{l|l|l|l}
    \toprule
    Training data & Fine-tuning data  &Testing data     & NLL $\downarrow$ \\\hline
    ZuCo~NR~\cite{hollenstein2018zuco}    &-         & ZuCo~NR~\cite{hollenstein2018zuco}    & 2.653 $\pm$ 0.020 \\
    CELER~L1~\cite{berzak2022celer}   &-         & ZuCo~NR~\cite{hollenstein2018zuco}    & 3.060 $\pm$ 0.026 \\
    CELER~L1~\cite{berzak2022celer}   &ZuCo~NR~\cite{hollenstein2018zuco}     & ZuCo~NR~\cite{hollenstein2018zuco}     &  \textbf{2.613 $\pm$ 0.019} \\
          
    \bottomrule
    \end{tabular}
\end{center}

\end{table*}

\subsection{Ablation Study}
\label{sec:ablation-study}
In this section, we investigate the inner workings of Eyettention by conducting an ablation study to examine the influence of each of its components and input features. An overview of the results is shown in Table~\ref{tab: ablation}. For both Chinese (BSC) and English (CELER~L1), we find that across all splits, (i) windowed attention (rather than standard global attention) and (ii) the Word-Sequence Encoder that provides the Decoder with a representation of the  stimulus sentence that maintains its original word order significantly contribute to the model's performance. 
Adding the word length information to the encoded stimulus further improves model performance in the English data across all splits, whereas for Chinese it does not have a significant impact on the model's performance. 
Neither for Chinese nor English, applying a Gaussian filter to the output of the windowed attention leads to a significant improvement in the model's performance compared to non-smoothed windowed attention. 
When we skew the Gaussian window to the right (D$_{left}$=1, D$_{right}$=2, $\sigma$=2D) to mimic the asymmetry of the perceptual span while reading, no significant improvements are observed, except for the New Reader split for Chinese. Results for other parameter combinations of the asymmetric Gaussian window are available in Table~\ref{tab: asym-ablation} of the Appendix. When applying only a Gaussian kernel without a local window, the model's performance is significantly worse than the full model on the English data, whereas it yields similar results as the full model or the one with only windowed attention on the Chinese data. 
The contribution of the fixational input features differs between the two languages. Whereas for English neither landing position nor fixation duration improve model performance, for Chinese, landing position significantly improves performance across all splits.

\begin{table*}[]
\small
\caption{\emph{Ablation Study}. We report NLL $\pm$ standard error for 5-fold CV/random resampling for the New Reader split, New Sentence split, and New Reader / New Sentence split, respectively.  
The dagger $\dagger$ indicates models significantly worse than the Eyettention model and the asterisk * indicates models significantly better than the Eyettention model. }

\label{tab: ablation}

\begin{center}
    \begin{tabular}{l|l|l|l|l}
    \toprule
    && \multicolumn{3}{c}{Train/test split (NLL$\downarrow$)}\\
     &                              &                   &                 & New Reader /           \\
      Data Set&  Model                            & New Sentence                  & New Reader                 &  New Sentence          \\\hline
        BSC   &Eyettention                         &1.856 $\pm$ 0.007              & 1.875 $\pm$ 0.007          & 1.840 $\pm$ 0.017 \\
    (Chinese) &w/o word length         & 1.858 $\pm$ 0.007             & 1.871 $\pm$ 0.007          & 1.855 $\pm$ 0.016 \\
              &w/o fixation duration   & 1.861 $\pm$ 0.007             & 1.881 $\pm$ 0.007          & 1.847 $\pm$ 0.017 \\
              &w/o landing position    & 1.936 $\pm$ 0.007$\dagger$    & 1.952 $\pm$ 0.007$\dagger$          & 1.927 $\pm$ 0.016$\dagger$  \\
              &w/o Gaussian kernel     & 1.846 $\pm$ 0.007    & \textbf{1.832 $\pm$ 0.007}*           &1.840 $\pm$ 0.017 \\
              &w/ right skewed Gaussian kernel
              &\textbf{1.838 $\pm$ 0.007} & 1.853 $\pm$ 0.007*             & \textbf{1.831 $\pm$ 0.017}         \\
              &w/o local window   &1.87 $\pm$ 0.007     & 1.891 $\pm$ 0.007          & 1.865 $\pm$ 0.017 \\
              &w/o local window, w/o Gaussian kernel     &1.901 $\pm$ 0.007$\dagger$     & 1.907 $\pm$ 0.008$\dagger$          & 1.900 $\pm$ 0.017$\dagger$  \\
              &w/o Word-Sequence Encoder           & 1.994 $\pm$ 0.007$\dagger$    & 2.005 $\pm$ 0.007$\dagger$          & 1.984 $\pm$ 0.016$\dagger$ \\\hline 
          
    CELER~L1   &Eyettention                      &\textbf{2.277 $\pm$ 0.005}       & \textbf{2.267 $\pm$ 0.005}         &2.297 $\pm$ 0.011\\
               (English) &w/o word length       & 2.375 $\pm$ 0.005$\dagger$      & 2.379 $\pm$ 0.005$\dagger$         &2.400 $\pm$ 0.010$\dagger$ \\
               &w/o fixation duration & 2.281 $\pm$ 0.005               & 2.276 $\pm$ 0.005                  &2.296 $\pm$ 0.011\\
               &w/o landing position  & 2.284 $\pm$ 0.006               & 2.271 $\pm$ 0.005                  &\textbf{2.292 $\pm$ 0.011}\\
               &w/o Gaussian kernel   &2.278 $\pm$ 0.005                & 2.271 $\pm$ 0.006                  &2.301 $\pm$ 0.011   \\
               &w/ right skewed Gaussian kernel  
               & 2.318 $\pm$ 0.005$\dagger$             & 2.319 $\pm$ 0.005$\dagger$          & 2.345 $\pm$ 0.011$\dagger$\\
               &w/o local window       &  2.356 $\pm$ 0.005$\dagger$               &  2.348 $\pm$ 0.005$\dagger$                  &2.379 $\pm$ 0.011$\dagger$    \\
               &w/o local window, w/o Gaussian kernel    & 2.344 $\pm$ 0.005$\dagger$      & 2.320  $\pm$ 0.006$\dagger$         &2.377 $\pm$ 0.011$\dagger$     \\
               &w/o Word-Sequence Encoder         & 2.607 $\pm$ 0.005$\dagger$      &  2.599 $\pm$ 0.005$\dagger$          &2.631 $\pm$ 0.01$\dagger$ \\
    \bottomrule 
    \end{tabular}

\end{center}

\end{table*}

\subsection{Model Inspection}
\label{sec:diff-saccades}
In this Section, we  provide a qualitative inspection of how our model computes the prediction for the next fixation location. Examples for generated scanpaths can be found in Appendix~\ref{sec:pred-vis}. We first look into the model's learned attention, and then examine the model's ability to predict saccades of different ranges (e.g. long and short saccades).

\begin{figure*}[ht]
\captionsetup[subfigure]{justification=centering}
    \subfloat[Eyettention\label{subfig:atten_map_local_wgau}]{%
      \includegraphics[trim=35 6 70 0,clip,width=0.23\textwidth,keepaspectratio]{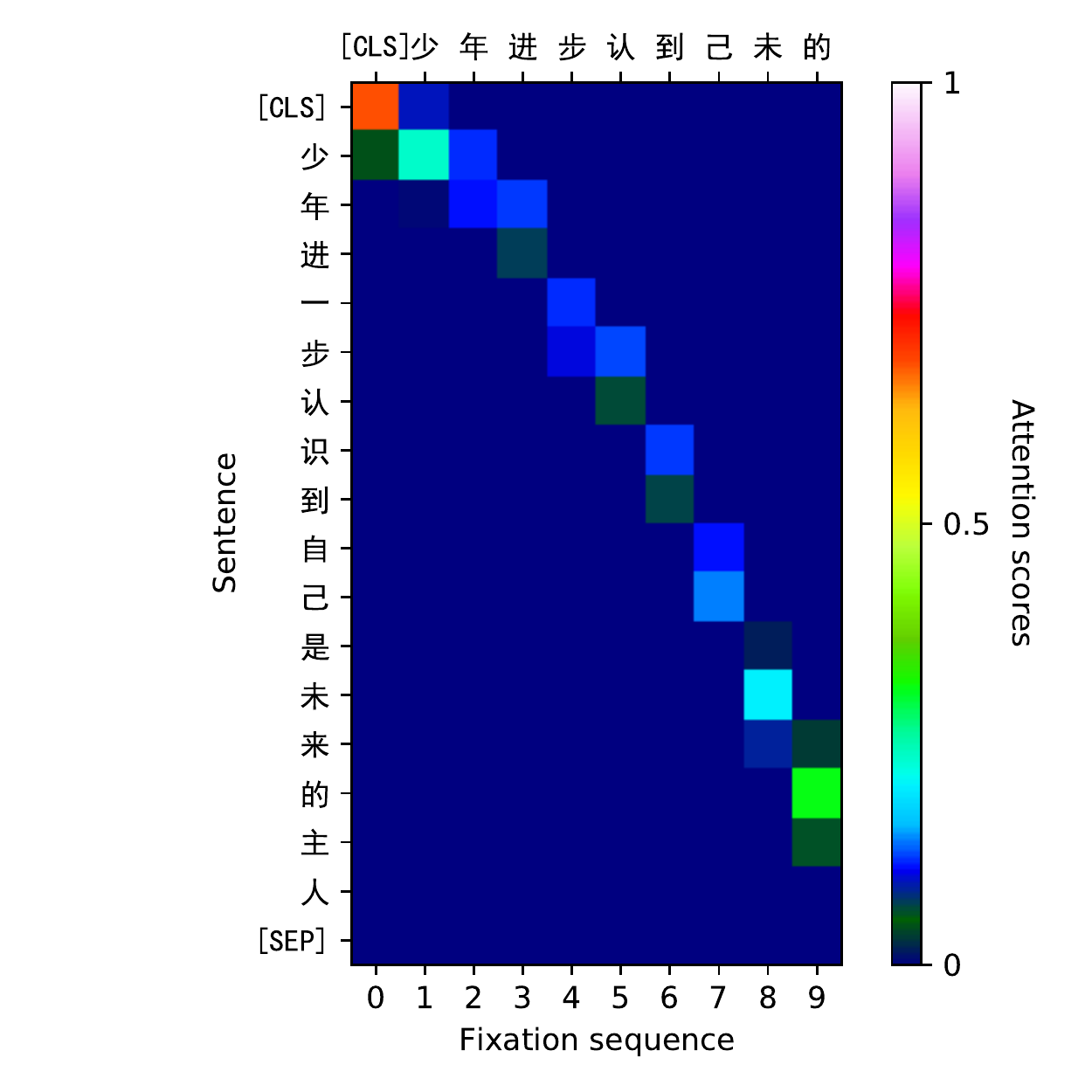}
    }
    \subfloat[Eyettention w/o Gaussian kernel\label{subfig:atten_map_local_wogau}]{%
      \includegraphics[trim=35 6 70 0,clip,width=0.23\textwidth,keepaspectratio]{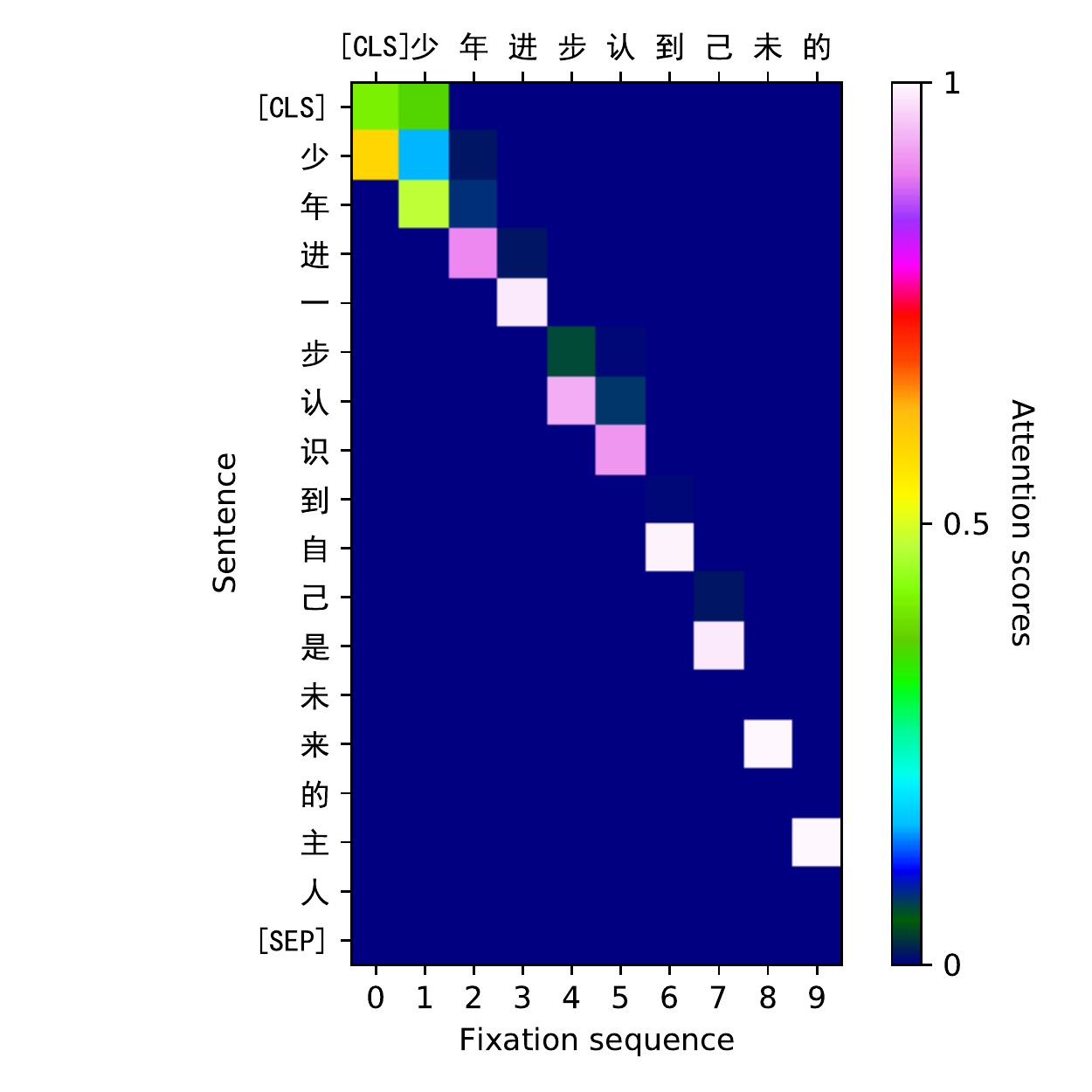}
    }
    \subfloat[Eyettention w/o local window\label{subfig:atten_map_global_wogau}]{%
      \includegraphics[trim=35 6 70 0,clip,width=0.23\textwidth,keepaspectratio]{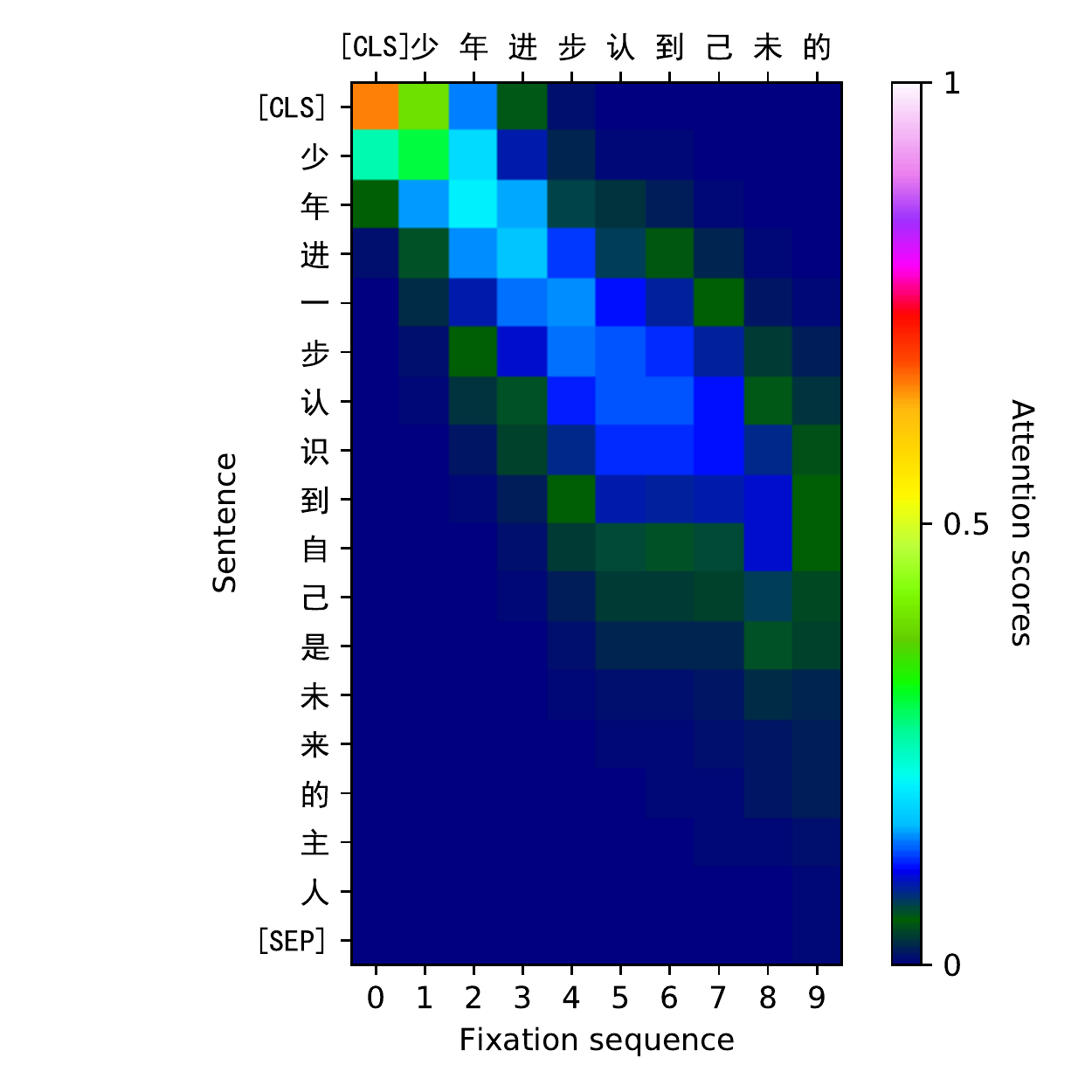}
    }
    \subfloat[Eyettention w/o local window, w/o Gaussian kernel\label{subfig:atten_map_global}]{%
      \includegraphics[trim=35 6 70 0,clip,width=0.23\textwidth,keepaspectratio]{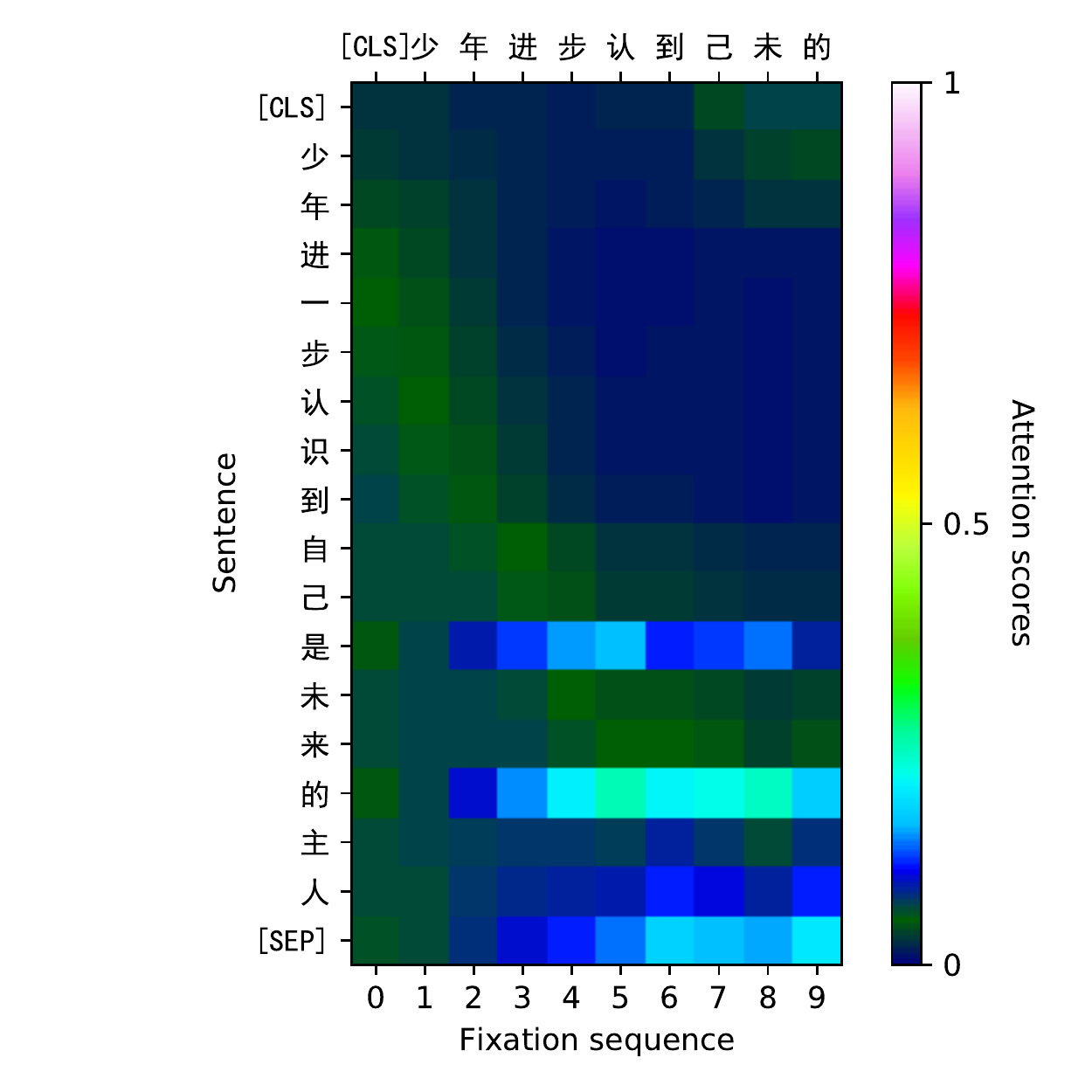}
    }

    \subfloat[Eyettention\label{subfig:atten_map_celer_local_wgau}]{%
      \includegraphics[trim=12 0 62 0,clip,width=0.23\textwidth,keepaspectratio]{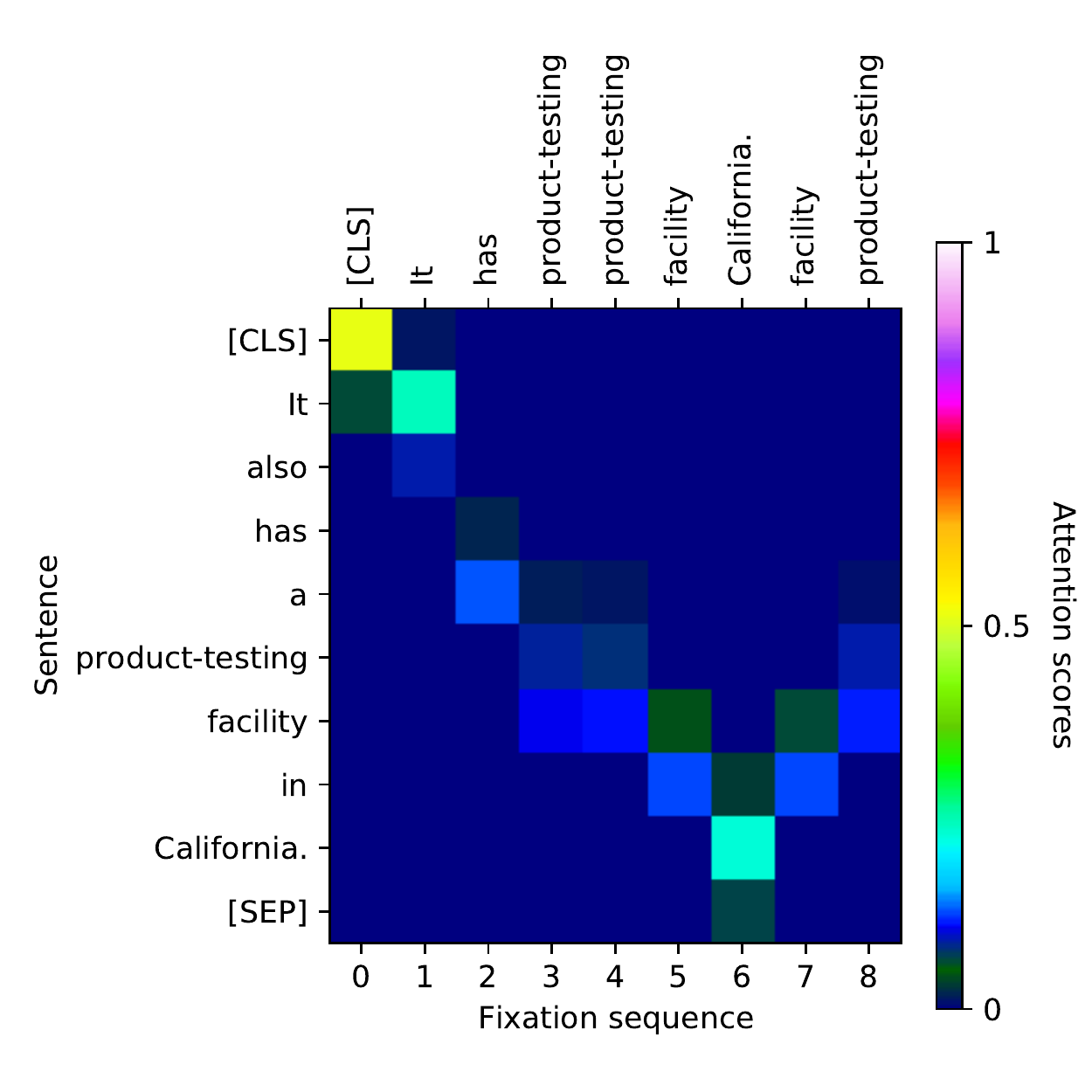}
    }
    \subfloat[Eyettention w/o Gaussian kernel\label{subfig:atten_map_celer_local_wogau}]{%
      \includegraphics[trim=12 0 62
      0,clip,width=0.23\textwidth,keepaspectratio]{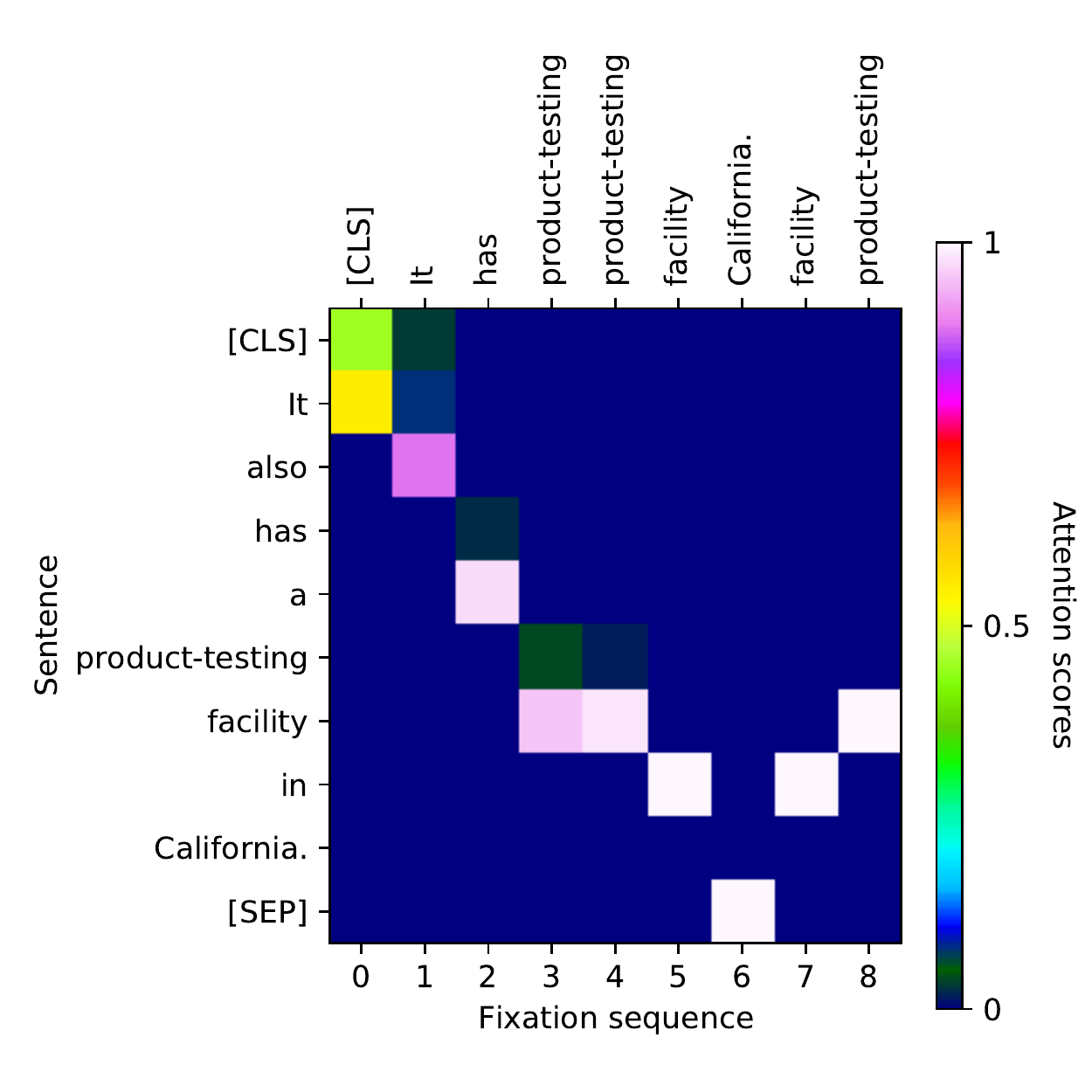}
    }
    \subfloat[Eyettention w/o local window\label{subfig:atten_map_celer_global_wogau}]{%
      \includegraphics[trim=12 0 62
      0,clip,width=0.23\textwidth,keepaspectratio]{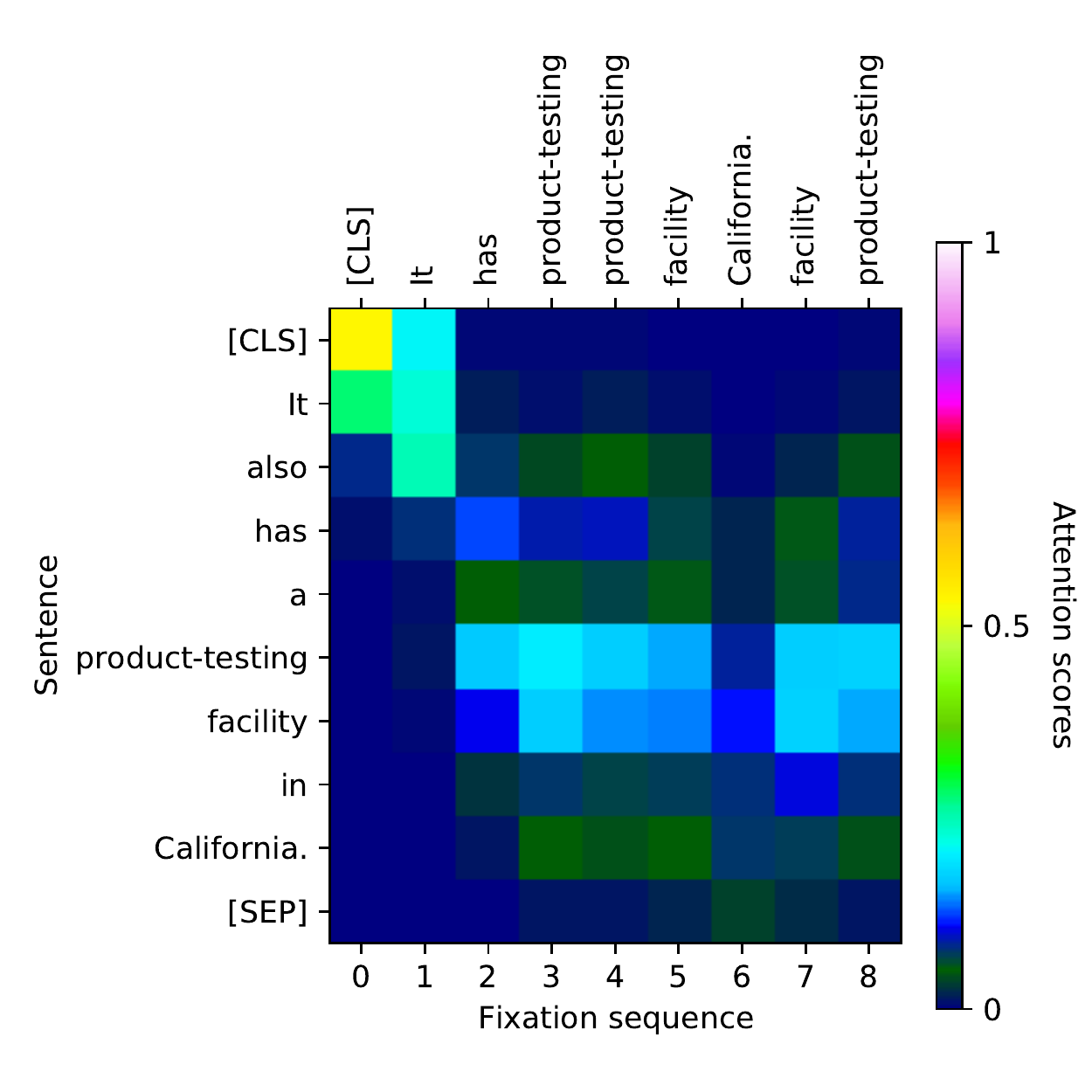}
    }
    \subfloat[Eyettention w/o local window, w/o Gaussian kernel\label{subfig:atten_map_celer_global}]{%
      \includegraphics[trim=12 0 62 0,clip,width=0.23\textwidth,keepaspectratio]{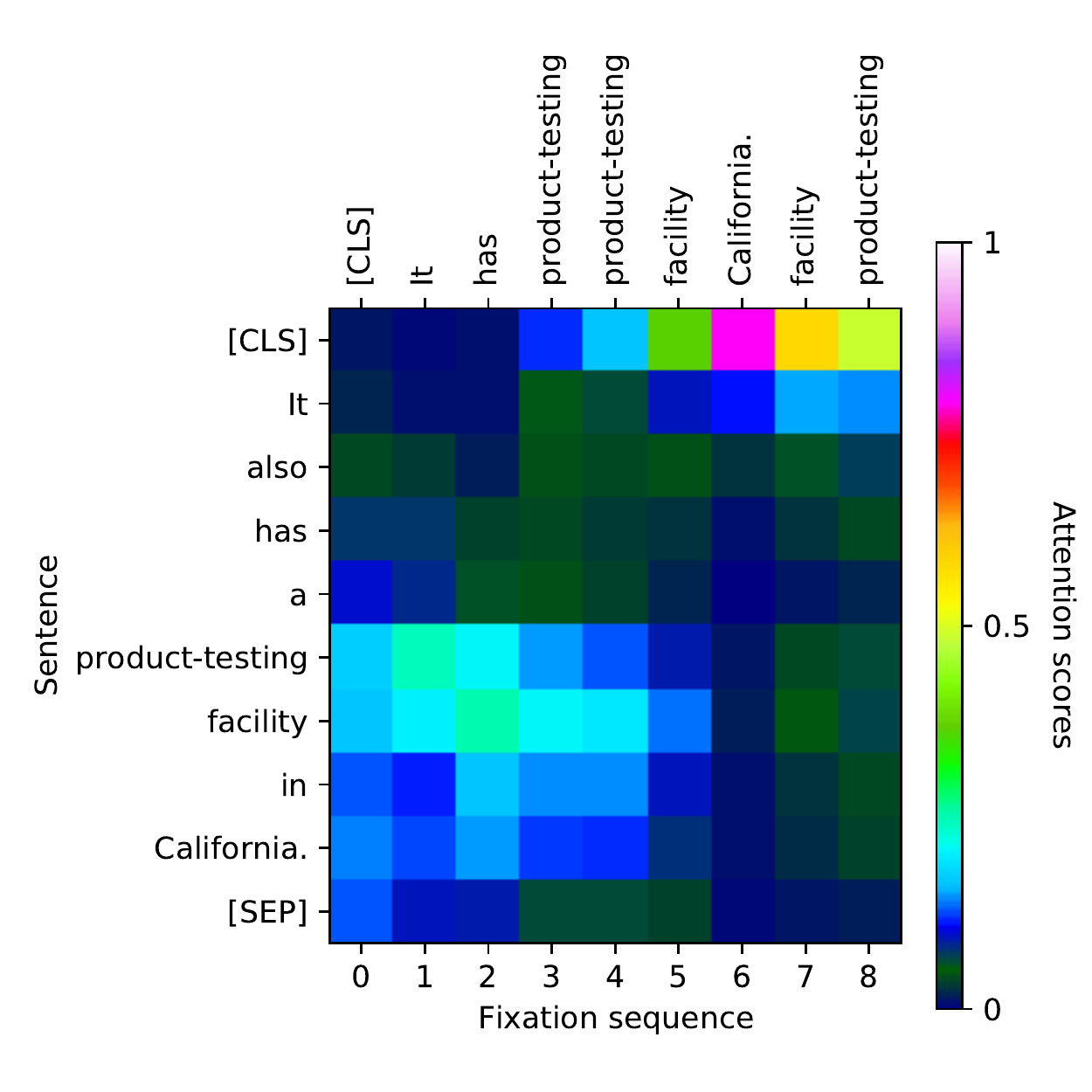}
    }
      \includegraphics[trim=300 0 5 0,clip,width=0.043\textwidth,keepaspectratio]{Figures/celer_batch1_idx95_global.pdf}
    
    \caption{Visualization of alignment scores for local attention with Gaussian kernel vs. local attention without Gaussian kernel vs. global attention with Gaussian kernel vs. global attention without Gaussian kernel for BSC data set (\ref{subfig:atten_map_local_wgau},~\ref{subfig:atten_map_local_wogau},~\ref{subfig:atten_map_global_wogau},~\ref{subfig:atten_map_global}) and CELER data set (\ref{subfig:atten_map_celer_local_wgau},~\ref{subfig:atten_map_celer_local_wogau},~\ref{subfig:atten_map_celer_global_wogau},~\ref{subfig:atten_map_celer_global}). The y-axis shows the complete stimulus sentence while the horizontal-axis shows the fixation sequence (bottom) together with the fixated words (top). }
    \label{fig:atten_map}

  \end{figure*}

\emph{Model Attention}. Figure~\ref{fig:atten_map} visualizes the model's attention for a Chinese and an English exemplary scanpath. The figure shows our proposed local attention mechanism (attention window with Gaussian kernel), local attention without Gaussian kernel, only a Gaussian kernel, and standard global attention (i.e., attending to the  entire sentence with neither a window nor a Gaussian kernel). We can see the clear advantage of local attention (with or without Gaussian kernel) over global attention: whereas local attention helps the model to focus on the words in the surroundings of the current fixation, the more uniform distribution of global attention reflects the model's uncertainty about what words are relevant for planning the next saccade's target. The Gaussian kernel smooths the attention around the current fixation with a tendency to a left-skew whereas the local attention window without Gaussian kernel tends to focus on the upcoming word. The attention by Eyettention w/o local window, i.e., using only a Gaussian kernel to mimic (para-)foveal vision, yields different patterns for English and Chinese. We see that for Chinese, the field of view is restricted to a local window whereas for the English corpus the attention shows similarities to global attention.

\begin{figure}[!ht]

    \subfloat[BSC data set, New Reader / New Sentence Split.\label{subfig:sac_type_BSC_NRS}]{%
  \includegraphics[width=0.49\textwidth,keepaspectratio]{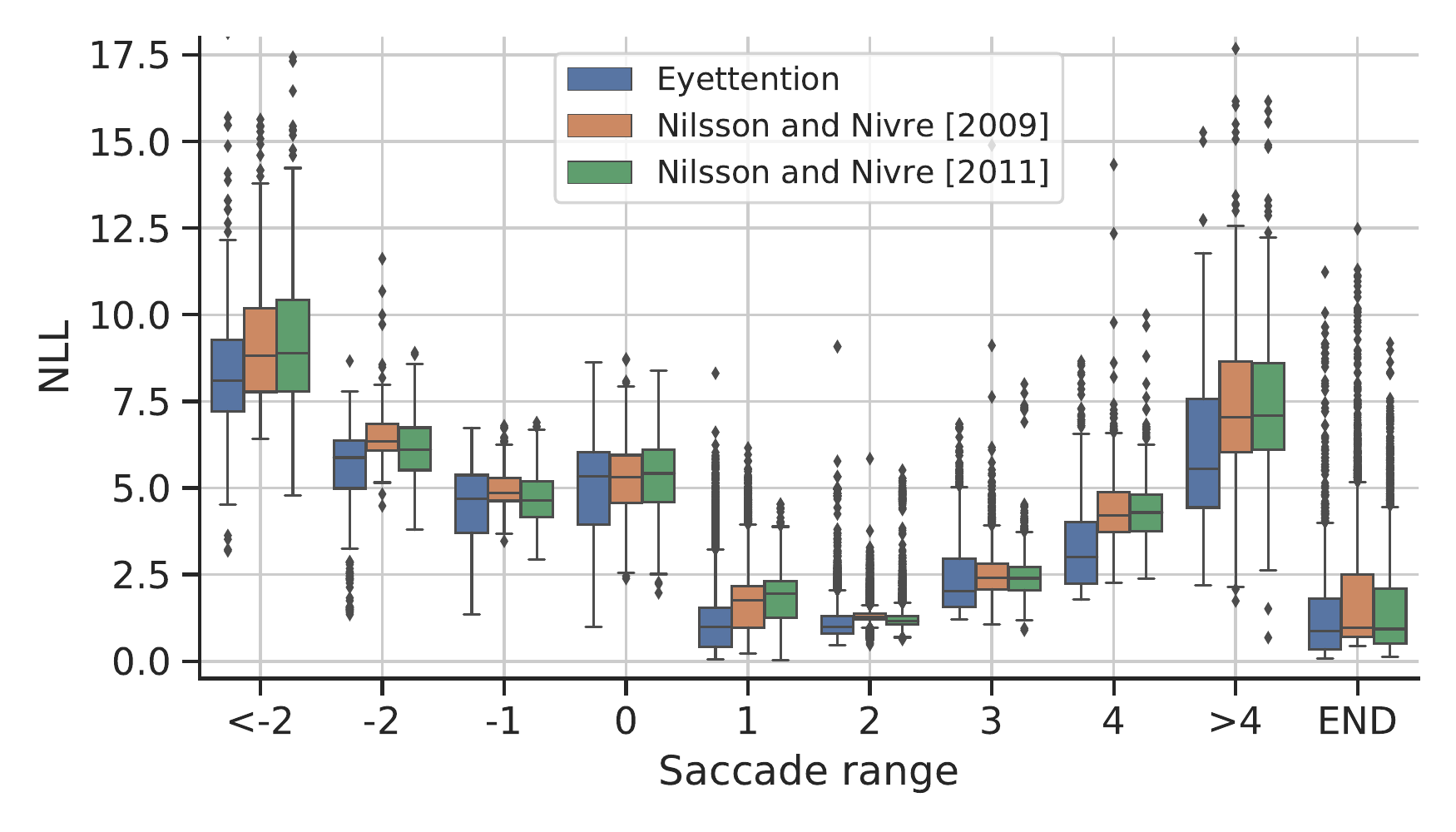}
    }
    \hfill
    \subfloat[CELER L1 data set, New Reader / New Sentence Split.\label{subfig:sac_type_BSC_celer}]{%
      \includegraphics[width=0.49\textwidth,keepaspectratio]{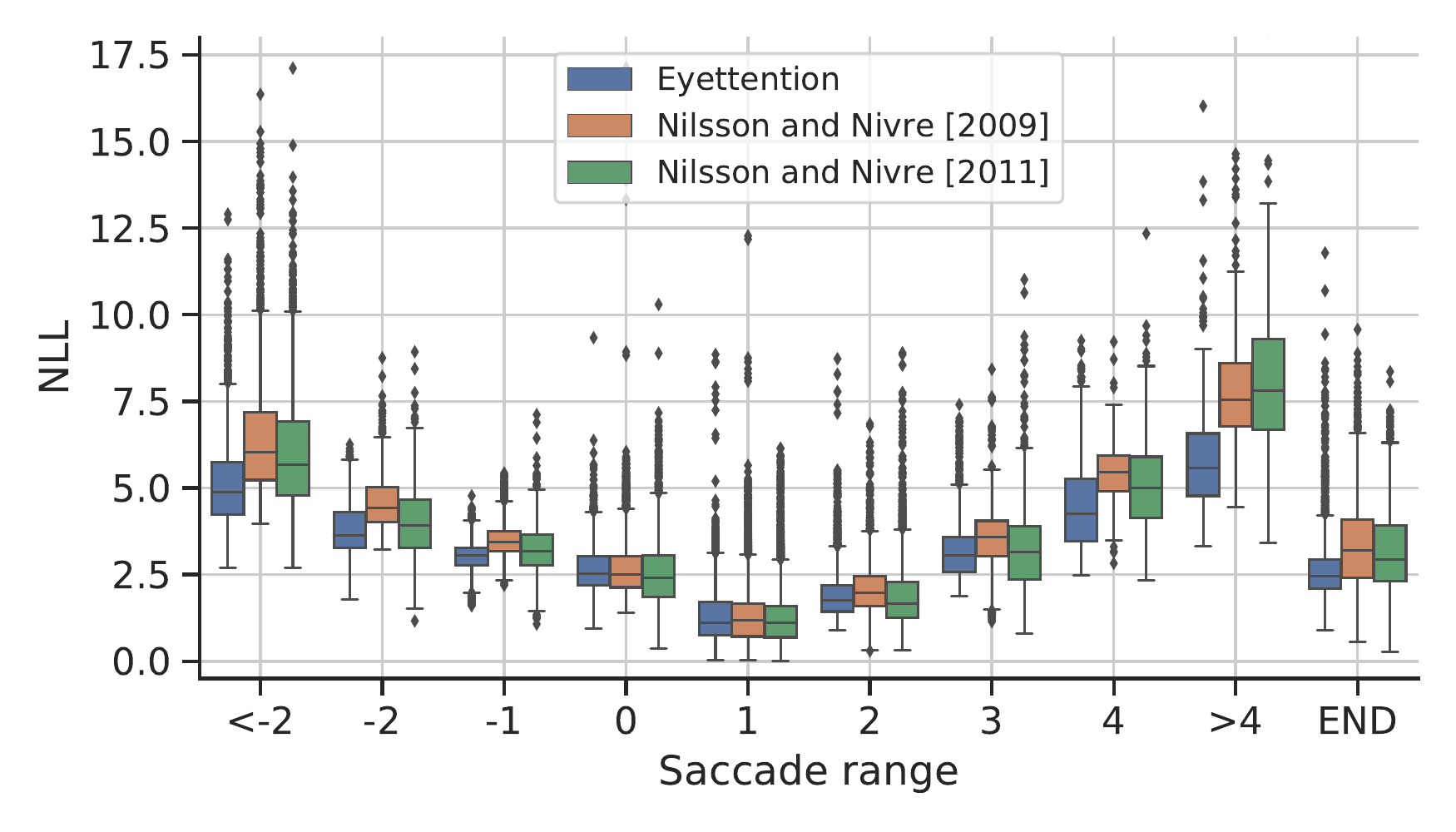}
    }
    \caption{NLL over different saccade ranges for the different methods on BSC and CELER L1 data set.}
    \label{fig: sac_type}

  \end{figure}

\emph{Performance wrt Saccade Length}. Figure~\ref{fig: sac_type} shows the performance of Eyettention together with those reference methods that allow to compute a NLL, as a function of the range of the next saccade. We focus here on the New Reader / New Sentence split, the results for the other splits are provided in Figure~\ref{fig: sac_type_rest} of the Appendix. We observe that overall, Eyettention outperforms the other models. Further, we find that all models perform best for short progressive saccades ranging from $1$ to $3$ words. For long progressive saccades ($>4$ words) and long regressions ($<-2$ words) the advantage of Eyettention over the other models is most pronounced. Overall, the performance on long regressions is better for the English data than for Chinese. Finally, the models are very good at predicting the end of a scanpath: their accuracy in predicting the end-of-scanpath token is on par with predicting short progressive saccades.


\section{Discussion}
\label{sec:discussion}
Our proposed neural-attention-based dual-sequence model \emph{Eyettention} outperforms the state-of-the-art in all of the investigated application scenarios represented by the different train/test splits, across metrics and languages/scripts. The key innovation of our approach leading to this performance gain---as demonstrated by the ablation study---is to incorporate the sequential nature of both, the stimulus sentence sorted according to the grammatical rules of the language (linguistic axis), and the chronologically sorted scanpath (temporal axis), into a single model architecture. We then align deep neural representations of the stimulus sentence and the scanpath up to the current position using local-Gaussian cross-attention to mimic the interplay between human visual field and human attention. 
The ablation study together with the qualitative inspection of the model's learned attention shows that our proposed combination of a local attention window with a Gaussian kernel centered around the current fixation---aimed at restricting the neural attention to the words in or around the fovea of the current fixation---(i) is critical for our model's performance and (ii) mimics human foveal and parafoveal vision quite well: the model attends to the  currently fixated word and, with reduced intensity, its immediate neighbors. First, our ablation study has demonstrated that helping the model attend to words around the currently fixated area of the stimulus sentence is crucial;
our proposed combination of a local attention window with a Gaussian kernel yields similar results compared to a local window only. Using only a Gaussian kernel also yields comparable results on the Chinese data set, while it is significantly worse than the other approaches on the English data. We hypothesize that this might arise from the larger variance in sentence length in the CELER data set. Qualitatively we note that when removing the Gaussian kernel and only using a local attention window, the model's attention is less smooth and only focuses on the word to the right of the currently fixated word.

Second, this entirely data-driven right-shift observed in the allocation of the model's attention weights aligns with the theoretically motivated property of SWIFT~\citep{engbert2005swift} that introduces a right-skewed spatial distribution of attention in the model. 
However, unlike the SWIFT model, there is no need for Eyettention to use a right-skewed Gaussian window to mimic the asymmetry of the human perceptual span during reading. Eyettention learns the right-shift from the data and encodes it in the attention weights (i.e., although the attention window is symmetric, the learned attention weights show the expected right-skew).

Instead of interpreting the local attention window in terms of perceptual span, we can also interpret it as an analogy of the local search space of the human language processor during the retrieval of items from working memory. In this interpretation, the fixation encoder queries locally available items (i.e., words) — the locality determined by the window size — from working memory, that are subsequently parsed by the language processor (decoder) to determine the next fixation location. Although Eyettention has not been designed as a cognitive model, its preference for a local attention window can be compared to humans' tendency to process sentences with a strong focus on local rather than global context, leading to phenomena such as \textit{late closure} \cite{frazier1979comprehending}, \textit{local coherence}~\citep{tabor2004evidence, smith2018toward} and shallow processing~\citep{ferreira2003misinterpretation,swets2008underspecification}.

Besides highlighting the necessity to locally constrain the model's attention, the ablation study has shown that using the landing positions as input feature significantly improves our model's performance. We hypothesize that the model learns to recognize whether a saccade overshot the target word, typically leading to a corrective micro-saccade. 

Although our approach outperforms the current state-of-the-art, the comparison with between-human scanpath similarity shows that Eyettention can still be improved upon. While outperforming the reference methods, the major weakness of Eyettention is clearly long regressive and progressive saccades. However, we see a considerably better result on the CELER data compared to the BSC data, which is presumably due to the longer sentences, and, consequently, higher number of long saccades, contained in the CELER data. This observation indicates that providing training data that includes long and complex sentences that trigger long regressions and progressions might remedy the model's difficulties with accurately predicting long saccades. 

The model's potential improvement as more training data becomes available is also evident in our cross-data set evaluation:  
Pre-training on a large data set, and then fine-tuning with only a few instances of the target data set with different stimulus presentation styles and hardware setups leads to very large improvements (cf.\, Figure~\ref{fig:finetune}b). 
The critical next step will be  to investigate whether the advantage of pre-training Eyettention also persists when fine-tuning for task-specific reading. This would allow us to generate task-specific synthetic data from just a few instances of human scanpaths, and close the gap to NLP-application-relevant scanpaths. In future research we want to further investigate how synthetic scanpaths generated with Eyettention can be leveraged to improve NLP downstream tasks. In addition, we are planning to extend our model to predict other fixation attributes 
to increase its applicability to a wide range of downstream tasks. Finally, it would be interesting to compare the BERT-derived features with the engineered features used in previous studies to analyze the impact of different linguistic representations on scanpath prediction.

\section{Conclusion}

\label{sec:conclusion}
We have presented Eyettention, the first end-to-end trained dual-sequence encoder-encoder model for next fixation prediction. Departing from earlier approaches, our model processes a given scanpath history and the presented text simultaneously and aligns the linguistic and the temporal sequence axes using a local cross-attention mechanism mimicking the human visual field. Eyettention not only outperforms both, cognitive and machine-learning based state-of-the-art models of eye movement control in reading, but also generalizes across data sets and even across typologically different languages with different scripts.

\begin{acks}
The authors would like to thank Maximilian Rabe for his support regarding the SWIFT model. This work was partially funded by the German Federal Ministry of Education and Research under grant 01$\vert$ S20043.
\end{acks}

\bibliographystyle{ACM-Reference-Format}
\bibliography{main.bib}


\begin{thebibliography}{57}


\ifx \showCODEN    \undefined \def \showCODEN     #1{\unskip}     \fi
\ifx \showDOI      \undefined \def \showDOI       #1{#1}\fi
\ifx \showISBNx    \undefined \def \showISBNx     #1{\unskip}     \fi
\ifx \showISBNxiii \undefined \def \showISBNxiii  #1{\unskip}     \fi
\ifx \showISSN     \undefined \def \showISSN      #1{\unskip}     \fi
\ifx \showLCCN     \undefined \def \showLCCN      #1{\unskip}     \fi
\ifx \shownote     \undefined \def \shownote      #1{#1}          \fi
\ifx \showarticletitle \undefined \def \showarticletitle #1{#1}   \fi
\ifx \showURL      \undefined \def \showURL       {\relax}        \fi
\providecommand\bibfield[2]{#2}
\providecommand\bibinfo[2]{#2}
\providecommand\natexlab[1]{#1}
\providecommand\showeprint[2][]{arXiv:#2}

\bibitem[Ahn et~al\mbox{.}(2020)]%
        {Ahn2020TowardsBehavior}
\bibfield{author}{\bibinfo{person}{Seoyoung Ahn}, \bibinfo{person}{Conor
  Kelton}, \bibinfo{person}{Aruna Balasubramanian}, {and} \bibinfo{person}{Greg
  Zelinsky}.} \bibinfo{year}{2020}\natexlab{}.
\newblock \showarticletitle{Towards predicting reading comprehension from gaze
  behavior}. In \bibinfo{booktitle}{\emph{Proceedings of the 2020 Symposium on
  Eye Tracking Research and Applications}}. \bibinfo{address}{Stuttgart,
  Germany}, \bibinfo{pages}{1--5}.
\newblock


\bibitem[Barrett et~al\mbox{.}(2018)]%
        {barrett2018unsupervised}
\bibfield{author}{\bibinfo{person}{Maria Barrett}, \bibinfo{person}{Ana~Valeria
  Gonz{\'a}lez-Gardu{\~n}o}, \bibinfo{person}{Lea Frermann}, {and}
  \bibinfo{person}{Anders S{\o}gaard}.} \bibinfo{year}{2018}\natexlab{}.
\newblock \showarticletitle{Unsupervised induction of linguistic categories
  with records of reading, speaking, and writing}. In
  \bibinfo{booktitle}{\emph{Proceedings of the 2018 Conference of the North
  American Chapter of the Association for Computational Linguistics: Human
  Language Technologies}}. \bibinfo{address}{New Orleans, Louisiana},
  \bibinfo{pages}{2028--2038}.
\newblock


\bibitem[Barrett et~al\mbox{.}(2016)]%
        {barrett2016cross}
\bibfield{author}{\bibinfo{person}{Maria Barrett}, \bibinfo{person}{Frank
  Keller}, {and} \bibinfo{person}{Anders S{\o}gaard}.}
  \bibinfo{year}{2016}\natexlab{}.
\newblock \showarticletitle{Cross-lingual transfer of correlations between
  parts of speech and gaze features}. In \bibinfo{booktitle}{\emph{Proceedings
  of the 26th International Conference on Computational Linguistics:
  {T}echnical Papers}}. \bibinfo{address}{Osaka, Japan},
  \bibinfo{pages}{1330--1339}.
\newblock


\bibitem[Berzak et~al\mbox{.}(2018)]%
        {Berzak2018Assessing}
\bibfield{author}{\bibinfo{person}{Yevgeni Berzak}, \bibinfo{person}{Boris
  Katz}, {and} \bibinfo{person}{Roger Levy}.} \bibinfo{year}{2018}\natexlab{}.
\newblock \showarticletitle{Assessing language proficiency from eye movements
  in reading}. In \bibinfo{booktitle}{\emph{Proceedings of the 17th Annual
  Conference of the North American Chapter of the Association for Computational
  Linguistics: Human Language Technologies}}. \bibinfo{address}{New Orleans,
  Louisiana}, \bibinfo{pages}{1986--1996}.
\newblock


\bibitem[Berzak et~al\mbox{.}(2022)]%
        {berzak2022celer}
\bibfield{author}{\bibinfo{person}{Yevgeni Berzak}, \bibinfo{person}{Chie
  Nakamura}, \bibinfo{person}{Amelia Smith}, \bibinfo{person}{Emily Weng},
  \bibinfo{person}{Boris Katz}, \bibinfo{person}{Suzanne Flynn}, {and}
  \bibinfo{person}{Roger Levy}.} \bibinfo{year}{2022}\natexlab{}.
\newblock \showarticletitle{{CELER: A 365-participant corpus of eye movements
  in L1 and L2 English reading}}.
\newblock \bibinfo{journal}{\emph{Open Mind}} (\bibinfo{year}{2022}),
  \bibinfo{pages}{1--10}.
\newblock


\bibitem[{COST Association}(2022)]%
        {cost}
\bibfield{author}{\bibinfo{person}{{COST Association}}.}
  \bibinfo{year}{2022}\natexlab{}.
\newblock \bibinfo{title}{{Enabling multilingual eye-tracking data collection
  for human and machine language processing research (MultiplEYE)}}.
\newblock \bibinfo{howpublished}{\url{https://www.cost.eu/actions/CA21131/}}.
\newblock
\newblock
\shownote{[Online; accessed 03-November-2022]}.


\bibitem[Cristino et~al\mbox{.}(2010)]%
        {cristino2010scanmatch}
\bibfield{author}{\bibinfo{person}{Filipe Cristino},
  \bibinfo{person}{Sebastiaan Math{\^o}t}, \bibinfo{person}{Jan Theeuwes},
  {and} \bibinfo{person}{Iain~D Gilchrist}.} \bibinfo{year}{2010}\natexlab{}.
\newblock \showarticletitle{ScanMatch: A novel method for comparing fixation
  sequences}.
\newblock \bibinfo{journal}{\emph{Behavior Research Methods}}
  \bibinfo{volume}{42}, \bibinfo{number}{3} (\bibinfo{year}{2010}),
  \bibinfo{pages}{692--700}.
\newblock


\bibitem[Deng et~al\mbox{.}(2022)]%
        {deng2022detection}
\bibfield{author}{\bibinfo{person}{Shuwen Deng}, \bibinfo{person}{Paul Prasse},
  \bibinfo{person}{David~R. Reich}, \bibinfo{person}{Sabine Dziemian},
  \bibinfo{person}{Maja Stegenwallner-Schütz}, \bibinfo{person}{Daniel
  Krakowczyk}, \bibinfo{person}{Silvia Makowski}, \bibinfo{person}{Nicolas
  Langer}, \bibinfo{person}{Tobias Scheffer}, {and} \bibinfo{person}{Lena~A.
  Jäger}.} \bibinfo{year}{2022}\natexlab{}.
\newblock \showarticletitle{Detection of {ADHD} based on eye movements during
  natural viewing}. In \bibinfo{booktitle}{\emph{Proceedings of the European
  Conference on Machine Learning and Knowledge Discovery in Databases}}.
  \bibinfo{publisher}{Springer}, \bibinfo{address}{Grenoble, France}.
\newblock


\bibitem[Devlin et~al\mbox{.}(2019)]%
        {devlin2018bert}
\bibfield{author}{\bibinfo{person}{Jacob Devlin}, \bibinfo{person}{Ming~Wei
  Chang}, \bibinfo{person}{Kenton Lee}, {and} \bibinfo{person}{Kristina
  Toutanova}.} \bibinfo{year}{2019}\natexlab{}.
\newblock \showarticletitle{{BERT: Pre-training of deep bidirectional
  transformers for language understanding}}. In
  \bibinfo{booktitle}{\emph{Proceedings of the Conference of the North American
  Chapter of the Association for Computational Linguistics: Human Language
  Technologies}}, Vol.~\bibinfo{volume}{1}. \bibinfo{address}{Minneapolis, MN,
  USA}, \bibinfo{pages}{4171--4186}.
\newblock


\bibitem[Engbert et~al\mbox{.}(2002)]%
        {engbert2002dynamical}
\bibfield{author}{\bibinfo{person}{Ralf Engbert}, \bibinfo{person}{Andr{\'e}
  Longtin}, {and} \bibinfo{person}{Reinhold Kliegl}.}
  \bibinfo{year}{2002}\natexlab{}.
\newblock \showarticletitle{A dynamical model of saccade generation in reading
  based on spatially distributed lexical processing}.
\newblock \bibinfo{journal}{\emph{Vision Research}} \bibinfo{volume}{42},
  \bibinfo{number}{5} (\bibinfo{year}{2002}), \bibinfo{pages}{621--636}.
\newblock


\bibitem[Engbert et~al\mbox{.}(2005)]%
        {engbert2005swift}
\bibfield{author}{\bibinfo{person}{Ralf Engbert}, \bibinfo{person}{Antje
  Nuthmann}, \bibinfo{person}{Eike~M Richter}, {and} \bibinfo{person}{Reinhold
  Kliegl}.} \bibinfo{year}{2005}\natexlab{}.
\newblock \showarticletitle{S{WIFT}: a dynamical model of saccade generation
  during reading.}
\newblock \bibinfo{journal}{\emph{Psychological Review}} \bibinfo{volume}{112},
  \bibinfo{number}{4} (\bibinfo{year}{2005}), \bibinfo{pages}{777}.
\newblock


\bibitem[Engelmann et~al\mbox{.}(2013)]%
        {engelmann2013framework}
\bibfield{author}{\bibinfo{person}{Felix Engelmann}, \bibinfo{person}{Shravan
  Vasishth}, \bibinfo{person}{Ralf Engbert}, {and} \bibinfo{person}{Reinhold
  Kliegl}.} \bibinfo{year}{2013}\natexlab{}.
\newblock \showarticletitle{A framework for modeling the interaction of
  syntactic processing and eye movement control}.
\newblock \bibinfo{journal}{\emph{Topics in Cognitive Science}}
  \bibinfo{volume}{5}, \bibinfo{number}{3} (\bibinfo{year}{2013}),
  \bibinfo{pages}{452--474}.
\newblock


\bibitem[Ferreira(2003)]%
        {ferreira2003misinterpretation}
\bibfield{author}{\bibinfo{person}{Fernanda Ferreira}.}
  \bibinfo{year}{2003}\natexlab{}.
\newblock \showarticletitle{The misinterpretation of noncanonical sentences}.
\newblock \bibinfo{journal}{\emph{Cognitive Psychology}} \bibinfo{volume}{47},
  \bibinfo{number}{2} (\bibinfo{year}{2003}), \bibinfo{pages}{164--203}.
\newblock


\bibitem[Frazier(1979)]%
        {frazier1979comprehending}
\bibfield{author}{\bibinfo{person}{Lyn Frazier}.}
  \bibinfo{year}{1979}\natexlab{}.
\newblock \bibinfo{booktitle}{\emph{On comprehending sentences: Syntactic
  parsing strategies.}}
\newblock \bibinfo{publisher}{University of Connecticut}.
\newblock


\bibitem[Gehring et~al\mbox{.}(2017)]%
        {gehring2017convolutional}
\bibfield{author}{\bibinfo{person}{Jonas Gehring}, \bibinfo{person}{Michael
  Auli}, \bibinfo{person}{David Grangier}, \bibinfo{person}{Denis Yarats},
  {and} \bibinfo{person}{Yann~N Dauphin}.} \bibinfo{year}{2017}\natexlab{}.
\newblock \showarticletitle{Convolutional sequence to sequence learning}. In
  \bibinfo{booktitle}{\emph{Proceedings of the 34th International Conference on
  Machine Learning - Volume 70}} \emph{(\bibinfo{series}{ICML'17})}.
  \bibinfo{address}{Sydney, Australia}, \bibinfo{pages}{1243--1252}.
\newblock


\bibitem[Gonz{\'a}lez-Gardu{\~n}o and S{\o}gaard(2017)]%
        {gonzalez-garduno-sogaard-2017}
\bibfield{author}{\bibinfo{person}{Ana~Valeria Gonz{\'a}lez-Gardu{\~n}o} {and}
  \bibinfo{person}{Anders S{\o}gaard}.} \bibinfo{year}{2017}\natexlab{}.
\newblock \showarticletitle{Using gaze to predict text readability}. In
  \bibinfo{booktitle}{\emph{Proceedings of the 12th Workshop on Innovative Use
  of {NLP} for Building Educational Applications, EMNLP}}.
  \bibinfo{address}{Copenhagen, Denmark}, \bibinfo{pages}{438--443}.
\newblock


\bibitem[Hahn and Keller(2016)]%
        {hahn2016modeling}
\bibfield{author}{\bibinfo{person}{Michael Hahn} {and} \bibinfo{person}{Frank
  Keller}.} \bibinfo{year}{2016}\natexlab{}.
\newblock \showarticletitle{Modeling human reading with neural attention}. In
  \bibinfo{booktitle}{\emph{Proceedings of the 2016 Conference on Empirical
  Methods in Natural Language Processing}}. \bibinfo{address}{Austin, Texas},
  \bibinfo{pages}{85--95}.
\newblock


\bibitem[Hahn and Keller(2023)]%
        {hahn2023modeling}
\bibfield{author}{\bibinfo{person}{Michael Hahn} {and} \bibinfo{person}{Frank
  Keller}.} \bibinfo{year}{2023}\natexlab{}.
\newblock \showarticletitle{Modeling task effects in human reading with neural
  network-based attention}.
\newblock \bibinfo{journal}{\emph{Cognition}}  \bibinfo{volume}{230}
  (\bibinfo{year}{2023}), \bibinfo{pages}{105289}.
\newblock


\bibitem[Haller et~al\mbox{.}(2022)]%
        {haller2022eye-tracking}
\bibfield{author}{\bibinfo{person}{Patrick Haller}, \bibinfo{person}{Andreas
  Säuberli}, \bibinfo{person}{Sarah~E. Kiener}, \bibinfo{person}{Jinger Pan},
  \bibinfo{person}{Ming Yan}, {and} \bibinfo{person}{Lena~A. Jäger}.}
  \bibinfo{year}{2022}\natexlab{}.
\newblock \showarticletitle{Eye-tracking based classification of Mandarin
  Chinese readers with and without dyslexia using neural sequence models}. In
  \bibinfo{booktitle}{\emph{Proceedings of the Workshop on Text Simplification,
  Accessibility, and Readability. EMNLP}}. \bibinfo{publisher}{Association for
  Computational Linguistics}, \bibinfo{address}{Abu Dhabi, UAE}.
\newblock


\bibitem[Hochreiter and Schmidhuber(1997)]%
        {hochreiter1997long}
\bibfield{author}{\bibinfo{person}{Sepp Hochreiter} {and}
  \bibinfo{person}{J{\"u}rgen Schmidhuber}.} \bibinfo{year}{1997}\natexlab{}.
\newblock \showarticletitle{Long short-term memory}.
\newblock \bibinfo{journal}{\emph{Neural Computation}} \bibinfo{volume}{9},
  \bibinfo{number}{8} (\bibinfo{year}{1997}), \bibinfo{pages}{1735--1780}.
\newblock


\bibitem[Hollenstein et~al\mbox{.}(2022)]%
        {hollenstein2022patterns}
\bibfield{author}{\bibinfo{person}{Nora Hollenstein}, \bibinfo{person}{Itziar
  Gonzalez-Dios}, \bibinfo{person}{Lisa Beinborn}, {and}
  \bibinfo{person}{Lena~A. Jäger}.} \bibinfo{year}{2022}\natexlab{}.
\newblock \showarticletitle{Patterns of text readability in human and predicted
  eye movements}. In \bibinfo{booktitle}{\emph{Proceedings of the Workshop on
  the Cognitive Aspects of the Lexicon, AACL}}. \bibinfo{address}{Online}.
\newblock


\bibitem[Hollenstein et~al\mbox{.}(2021)]%
        {hollenstein2021multilingual}
\bibfield{author}{\bibinfo{person}{Nora Hollenstein}, \bibinfo{person}{Federico
  Pirovano}, \bibinfo{person}{Ce Zhang}, \bibinfo{person}{Lena J{\"a}ger},
  {and} \bibinfo{person}{Lisa Beinborn}.} \bibinfo{year}{2021}\natexlab{}.
\newblock \showarticletitle{Multilingual Language Models Predict Human Reading
  Behavior}. In \bibinfo{booktitle}{\emph{Proceedings of the 2021 Conference of
  the North American Chapter of the Association for Computational Linguistics:
  Human Language Technologies}}. \bibinfo{address}{Online},
  \bibinfo{pages}{106--123}.
\newblock


\bibitem[Hollenstein et~al\mbox{.}(2018)]%
        {hollenstein2018zuco}
\bibfield{author}{\bibinfo{person}{Nora Hollenstein}, \bibinfo{person}{Jonathan
  Rotsztejn}, \bibinfo{person}{Marius Troendle}, \bibinfo{person}{Andreas
  Pedroni}, \bibinfo{person}{Ce Zhang}, {and} \bibinfo{person}{Nicolas
  Langer}.} \bibinfo{year}{2018}\natexlab{}.
\newblock \showarticletitle{ZuCo, a simultaneous {EEG} and eye-tracking
  resource for natural sentence reading}.
\newblock \bibinfo{journal}{\emph{Scientific Data}}  \bibinfo{volume}{5}
  (\bibinfo{year}{2018}), \bibinfo{pages}{180291}.
\newblock


\bibitem[Hollenstein et~al\mbox{.}(2020)]%
        {hollenstein2019zuco2}
\bibfield{author}{\bibinfo{person}{Nora Hollenstein}, \bibinfo{person}{Marius
  Troendle}, \bibinfo{person}{Ce Zhang}, {and} \bibinfo{person}{Nicolas
  Langer}.} \bibinfo{year}{2020}\natexlab{}.
\newblock \showarticletitle{{Z}u{C}o 2.0: A dataset of physiological recordings
  during natural reading and Annotation}. In
  \bibinfo{booktitle}{\emph{Proceedings of the Twelfth Language Resources and
  Evaluation Conference}}. \bibinfo{address}{Marseille, France},
  \bibinfo{pages}{138--146}.
\newblock


\bibitem[Hollenstein and Zhang(2019)]%
        {hollenstein-zhang-2019-entity}
\bibfield{author}{\bibinfo{person}{Nora Hollenstein} {and} \bibinfo{person}{Ce
  Zhang}.} \bibinfo{year}{2019}\natexlab{}.
\newblock \showarticletitle{Entity recognition at first sight: {I}mproving
  {NER} with eye movement information}. In
  \bibinfo{booktitle}{\emph{Proceedings of the 2019 Conference of the North
  {A}merican Chapter of the Association for Computational Linguistics: Human
  Language Technologies}}. \bibinfo{address}{Minneapolis, Minnesota},
  \bibinfo{pages}{1--10}.
\newblock


\bibitem[Jarodzka et~al\mbox{.}(2010)]%
        {jarodzka2010vector}
\bibfield{author}{\bibinfo{person}{Halszka Jarodzka}, \bibinfo{person}{Kenneth
  Holmqvist}, {and} \bibinfo{person}{Marcus Nystr{\"o}m}.}
  \bibinfo{year}{2010}\natexlab{}.
\newblock \showarticletitle{A vector-based, multidimensional scanpath
  similarity measure}. In \bibinfo{booktitle}{\emph{Proceedings of the 2010
  Symposium on Eye-Tracking Research and Applications}}
  \emph{(\bibinfo{series}{ETRA '10})}. \bibinfo{address}{Austin, Texas},
  \bibinfo{pages}{211–218}.
\newblock


\bibitem[Kingma and Ba(2015)]%
        {kingma2014adam}
\bibfield{author}{\bibinfo{person}{Diederik~P. Kingma} {and}
  \bibinfo{person}{Jimmy Ba}.} \bibinfo{year}{2015}\natexlab{}.
\newblock \showarticletitle{Adam: {A} Method for Stochastic Optimization}. In
  \bibinfo{booktitle}{\emph{Proceedings of the 3rd International Conference on
  Learning Representations, {ICLR}, San Diego, CA, USA, May 7-9, 2015}}.
\newblock


\bibitem[Klerke et~al\mbox{.}(2016)]%
        {klerke2016}
\bibfield{author}{\bibinfo{person}{Sigrid Klerke}, \bibinfo{person}{Yoav
  Goldberg}, {and} \bibinfo{person}{Anders S{\o}gaard}.}
  \bibinfo{year}{2016}\natexlab{}.
\newblock \showarticletitle{Improving sentence compression by learning to
  predict gaze}. In \bibinfo{booktitle}{\emph{Proceedings of the 2016
  Conference of the North {A}merican Chapter of the Association for
  Computational Linguistics: Human Language Technologies}}.
  \bibinfo{address}{San Diego, California}, \bibinfo{pages}{1528--1533}.
\newblock


\bibitem[K{\"u}mmerer and Bethge(2021)]%
        {kummerer2021state}
\bibfield{author}{\bibinfo{person}{Matthias K{\"u}mmerer} {and}
  \bibinfo{person}{Matthias Bethge}.} \bibinfo{year}{2021}\natexlab{}.
\newblock \showarticletitle{State-of-the-art in human scanpath prediction}.
\newblock \bibinfo{journal}{\emph{arXiv preprint arXiv:2102.12239}}
  (\bibinfo{year}{2021}).
\newblock


\bibitem[Levenshtein et~al\mbox{.}(1966)]%
        {levenshtein1966binary}
\bibfield{author}{\bibinfo{person}{Vladimir~I Levenshtein} {et~al\mbox{.}}}
  \bibinfo{year}{1966}\natexlab{}.
\newblock \showarticletitle{Binary codes capable of correcting deletions,
  insertions, and reversals}. In \bibinfo{booktitle}{\emph{Soviet Physics
  Doklady}}, Vol.~\bibinfo{volume}{10}. Soviet Union,
  \bibinfo{pages}{707--710}.
\newblock


\bibitem[Lim et~al\mbox{.}(2020)]%
        {lim2020emotion}
\bibfield{author}{\bibinfo{person}{Jia~Zheng Lim}, \bibinfo{person}{James
  Mountstephens}, {and} \bibinfo{person}{Jason Teo}.}
  \bibinfo{year}{2020}\natexlab{}.
\newblock \showarticletitle{Emotion recognition using eye-tracking: taxonomy,
  review and current challenges}.
\newblock \bibinfo{journal}{\emph{Sensors}} \bibinfo{volume}{20},
  \bibinfo{number}{8} (\bibinfo{year}{2020}), \bibinfo{pages}{2384}.
\newblock


\bibitem[Luong et~al\mbox{.}(2015)]%
        {luong2015effective}
\bibfield{author}{\bibinfo{person}{Thang Luong}, \bibinfo{person}{Hieu Pham},
  {and} \bibinfo{person}{Christopher~D. Manning}.}
  \bibinfo{year}{2015}\natexlab{}.
\newblock \showarticletitle{Effective Approaches to Attention-based Neural
  Machine Translation}. In \bibinfo{booktitle}{\emph{Proceedings of the 2015
  Conference on Empirical Methods in Natural Language Processing}}.
  \bibinfo{address}{Lisbon, Portugal}, \bibinfo{pages}{1412--1421}.
\newblock


\bibitem[Merkx and Frank(2021)]%
        {merkx-frank-2021-human}
\bibfield{author}{\bibinfo{person}{Danny Merkx} {and}
  \bibinfo{person}{Stefan~L. Frank}.} \bibinfo{year}{2021}\natexlab{}.
\newblock \showarticletitle{Human sentence processing: recurrence or
  attention?}. In \bibinfo{booktitle}{\emph{Proceedings of the Workshop on
  Cognitive Modeling and Computational Linguistics}}.
  \bibinfo{address}{Online}, \bibinfo{pages}{12--22}.
\newblock


\bibitem[Mishra et~al\mbox{.}(2016)]%
        {mishra2017leveraging}
\bibfield{author}{\bibinfo{person}{Abhijit Mishra}, \bibinfo{person}{Diptesh
  Kanojia}, \bibinfo{person}{Seema Nagar}, \bibinfo{person}{Kuntal Dey}, {and}
  \bibinfo{person}{Pushpak Bhattacharyya}.} \bibinfo{year}{2016}\natexlab{}.
\newblock \showarticletitle{Leveraging Cognitive Features for Sentiment
  Analysis}. In \bibinfo{booktitle}{\emph{Proceedings of the 20th {SIGNLL}
  Conference on Computational Natural Language Learning}}.
  \bibinfo{address}{Berlin, Germany}, \bibinfo{pages}{156--166}.
\newblock


\bibitem[Nilsson and Nivre(2009)]%
        {nilsson2009learning}
\bibfield{author}{\bibinfo{person}{Mattias Nilsson} {and}
  \bibinfo{person}{Joakim Nivre}.} \bibinfo{year}{2009}\natexlab{}.
\newblock \showarticletitle{Learning where to look: Modeling eye movements in
  reading}. In \bibinfo{booktitle}{\emph{Proceedings of the Thirteenth
  Conference on Computational Natural Language Learning}}.
  \bibinfo{address}{Boulder, Colorado}, \bibinfo{pages}{93--101}.
\newblock


\bibitem[Nilsson and Nivre(2010)]%
        {nilsson2010towards}
\bibfield{author}{\bibinfo{person}{Mattias Nilsson} {and}
  \bibinfo{person}{Joakim Nivre}.} \bibinfo{year}{2010}\natexlab{}.
\newblock \showarticletitle{Towards a data-driven model of eye movement control
  in reading}. In \bibinfo{booktitle}{\emph{Proceedings of the 2010 Workshop on
  Cognitive Modeling and Computational Linguistics, ACL}}.
  \bibinfo{address}{Uppsala, Sweden}, \bibinfo{pages}{63--71}.
\newblock


\bibitem[Nilsson and Nivre(2011)]%
        {nilsson2011entropy}
\bibfield{author}{\bibinfo{person}{Mattias Nilsson} {and}
  \bibinfo{person}{Joakim Nivre}.} \bibinfo{year}{2011}\natexlab{}.
\newblock \showarticletitle{Entropy-driven evaluation of models of eye movement
  control in reading}. In \bibinfo{booktitle}{\emph{Proceedings of the 8th
  International NLPCS Workshop}}. \bibinfo{address}{Copenhagen, Denmark},
  \bibinfo{pages}{201--212}.
\newblock


\bibitem[Pan et~al\mbox{.}(2021)]%
        {pan2021bsc}
\bibfield{author}{\bibinfo{person}{Jinger Pan}, \bibinfo{person}{Ming Yan},
  \bibinfo{person}{Eike~M. Richter}, \bibinfo{person}{Hua Shu}, {and}
  \bibinfo{person}{Reinhold Kliegl}.} \bibinfo{year}{2021}\natexlab{}.
\newblock \showarticletitle{The {B}eijing {S}entence {C}orpus: A {C}hinese
  sentence corpus with eye movement data and predictability norms}.
\newblock \bibinfo{journal}{\emph{Behavior Research Methods}}
  \bibinfo{volume}{2021} (\bibinfo{year}{2021}).
\newblock


\bibitem[Paszke et~al\mbox{.}(2019)]%
        {pytorch2019paszke}
\bibfield{author}{\bibinfo{person}{Adam Paszke}, \bibinfo{person}{Sam Gross},
  \bibinfo{person}{Francisco Massa}, \bibinfo{person}{Adam Lerer},
  \bibinfo{person}{James Bradbury}, \bibinfo{person}{Gregory Chanan},
  \bibinfo{person}{Trevor Killeen}, \bibinfo{person}{Zeming Lin},
  \bibinfo{person}{Natalia Gimelshein}, \bibinfo{person}{Luca Antiga},
  \bibinfo{person}{Alban Desmaison}, \bibinfo{person}{Andreas Kopf},
  \bibinfo{person}{Edward Yang}, \bibinfo{person}{Zachary DeVito},
  \bibinfo{person}{Martin Raison}, \bibinfo{person}{Alykhan Tejani},
  \bibinfo{person}{Sasank Chilamkurthy}, \bibinfo{person}{Benoit Steiner},
  \bibinfo{person}{Lu Fang}, \bibinfo{person}{Junjie Bai}, {and}
  \bibinfo{person}{Soumith Chintala}.} \bibinfo{year}{2019}\natexlab{}.
\newblock \showarticletitle{PyTorch: An Imperative Style, High-Performance Deep
  Learning Library}.
\newblock In \bibinfo{booktitle}{\emph{Proceedings of the 33rd International
  Conference on Neural Information Processing Systems}}.
  \bibinfo{pages}{8024--8035}.
\newblock


\bibitem[Raatikainen et~al\mbox{.}(2021)]%
        {Raatikainen2021DetectionData}
\bibfield{author}{\bibinfo{person}{Peter Raatikainen}, \bibinfo{person}{Jarkko
  Hautala}, \bibinfo{person}{Otto Loberg}, \bibinfo{person}{Tommi
  K{\"{a}}rkk{\"{a}}inen}, \bibinfo{person}{Paavo Lepp{\"{a}}nen}, {and}
  \bibinfo{person}{Paavo Nieminen}.} \bibinfo{year}{2021}\natexlab{}.
\newblock \showarticletitle{{Detection of developmental dyslexia with machine
  learning using eye movement data}}.
\newblock \bibinfo{journal}{\emph{Array}}  \bibinfo{volume}{12}
  (\bibinfo{year}{2021}), \bibinfo{pages}{100087}.
\newblock


\bibitem[Rabe et~al\mbox{.}(2021)]%
        {rabe2021bayes}
\bibfield{author}{\bibinfo{person}{Maximilian~M. Rabe}, \bibinfo{person}{Johan
  Chandra}, \bibinfo{person}{André Kr\"{u}gel},
  \bibinfo{person}{Stefan~Alexander Seelig}, \bibinfo{person}{Shravan
  Vasishth}, {and} \bibinfo{person}{Ralf Engbert}.}
  \bibinfo{year}{2021}\natexlab{}.
\newblock \showarticletitle{A Bayesian approach to dynamical modeling of
  eye-movement control in reading of normal, mirrored, and scrambled texts.}
\newblock \bibinfo{journal}{\emph{Psychol Rev}}  \bibinfo{volume}{128}
  (\bibinfo{year}{2021}), \bibinfo{pages}{803--823}.
\newblock
\urldef\tempurl%
\url{https://doi.org/doi:10.1037/rev0000268}
\showDOI{\tempurl}


\bibitem[Rayner(1998)]%
        {Rayner1998}
\bibfield{author}{\bibinfo{person}{Keith Rayner}.}
  \bibinfo{year}{1998}\natexlab{}.
\newblock \showarticletitle{Eye movements in reading and information
  processing: 20 years of research}.
\newblock \bibinfo{journal}{\emph{Psychological Bulletin}}
  \bibinfo{volume}{124}, \bibinfo{number}{3} (\bibinfo{year}{1998}),
  \bibinfo{pages}{372--422}.
\newblock


\bibitem[Rayner(2009)]%
        {rayner2009}
\bibfield{author}{\bibinfo{person}{Keith Rayner}.}
  \bibinfo{year}{2009}\natexlab{}.
\newblock \showarticletitle{The 35th {S}ir {F}rederick {B}artlett {L}ecture:
  Eye movements and attention in reading, scene perception, and visual search}.
\newblock \bibinfo{journal}{\emph{Quarterly Journal of Experimental
  Psychology}} \bibinfo{volume}{62}, \bibinfo{number}{8}
  (\bibinfo{year}{2009}), \bibinfo{pages}{1457--1506}.
\newblock


\bibitem[Rayner et~al\mbox{.}(2007)]%
        {rayner2007chineseezreader}
\bibfield{author}{\bibinfo{person}{Keith Rayner}, \bibinfo{person}{Xingshan
  Li}, {and} \bibinfo{person}{Alexander Pollatsek}.}
  \bibinfo{year}{2007}\natexlab{}.
\newblock \showarticletitle{Extending the E-Z Reader Model of Eye Movement
  Control to Chinese Readers}.
\newblock \bibinfo{journal}{\emph{Cognitive Science}} \bibinfo{volume}{31},
  \bibinfo{number}{6} (\bibinfo{year}{2007}), \bibinfo{pages}{1021--1033}.
\newblock
\urldef\tempurl%
\url{https://doi.org/10.1080/03640210701703824}
\showDOI{\tempurl}
\showeprint{https://onlinelibrary.wiley.com/doi/pdf/10.1080/03640210701703824}


\bibitem[Reich et~al\mbox{.}(2022)]%
        {reich2022inferring}
\bibfield{author}{\bibinfo{person}{David~Robert Reich}, \bibinfo{person}{Paul
  Prasse}, \bibinfo{person}{Chiara Tschirner}, \bibinfo{person}{Patrick
  Haller}, \bibinfo{person}{Frank Goldhammer}, {and} \bibinfo{person}{Lena~A.
  J\"{a}ger}.} \bibinfo{year}{2022}\natexlab{}.
\newblock \showarticletitle{Inferring Native and Non-Native Human Reading
  Comprehension and Subjective Text Difficulty from Scanpaths in Reading}. In
  \bibinfo{booktitle}{\emph{Proceedings of the 2022 Symposium on Eye Tracking
  Research and Applications}} (Seattle, WA, USA) \emph{(\bibinfo{series}{ETRA
  '22})}. \bibinfo{address}{Seattle, WA, USA}, Article \bibinfo{articleno}{23},
  \bibinfo{numpages}{8}~pages.
\newblock


\bibitem[Reichle et~al\mbox{.}(2003)]%
        {reichle2003ezreader}
\bibfield{author}{\bibinfo{person}{ED Reichle}, \bibinfo{person}{K Rayner},
  {and} \bibinfo{person}{Pollatsek A}.} \bibinfo{year}{2003}\natexlab{}.
\newblock \showarticletitle{The {E}-{Z} reader model of eye-movement control in
  reading: comparisons to other models}.
\newblock \bibinfo{journal}{\emph{The Behavioral and Brain Sciences}}
  \bibinfo{volume}{26} (\bibinfo{year}{2003}), \bibinfo{pages}{445--526}.
\newblock
Issue 4.


\bibitem[Siegelman et~al\mbox{.}(2022)]%
        {siegelman2022expanding}
\bibfield{author}{\bibinfo{person}{Noam Siegelman}, \bibinfo{person}{Sascha
  Schroeder}, \bibinfo{person}{Cengiz Acart{\"u}rk}, \bibinfo{person}{Hee-Don
  Ahn}, \bibinfo{person}{Svetlana Alexeeva}, \bibinfo{person}{Simona Amenta},
  \bibinfo{person}{Raymond Bertram}, \bibinfo{person}{Rolando Bonandrini},
  \bibinfo{person}{Marc Brysbaert}, \bibinfo{person}{Daria Chernova},
  {et~al\mbox{.}}} \bibinfo{year}{2022}\natexlab{}.
\newblock \showarticletitle{Expanding horizons of cross-linguistic research on
  reading: The Multilingual Eye-movement Corpus ({MECO})}.
\newblock \bibinfo{journal}{\emph{Behavior Research Methods}}
  (\bibinfo{year}{2022}), \bibinfo{pages}{1--21}.
\newblock


\bibitem[Smith and Tabor(2018)]%
        {smith2018toward}
\bibfield{author}{\bibinfo{person}{Garrett Smith} {and}
  \bibinfo{person}{Whitney Tabor}.} \bibinfo{year}{2018}\natexlab{}.
\newblock \showarticletitle{Toward a theory of timing effects in self-organized
  sentence processing}. In \bibinfo{booktitle}{\emph{Proceedings of the 16th
  International Conference on Cognitive Modeling}}. University of Wisconsin
  Madison, WI, \bibinfo{pages}{138--143}.
\newblock


\bibitem[Sood et~al\mbox{.}(2021)]%
        {sood2021multimodal}
\bibfield{author}{\bibinfo{person}{Ekta Sood}, \bibinfo{person}{Fabian
  K{\"{o}}gel}, \bibinfo{person}{Philipp M{\"{u}}ller},
  \bibinfo{person}{Dominike Thomas}, \bibinfo{person}{Mihai Bace}, {and}
  \bibinfo{person}{Andreas Bulling}.} \bibinfo{year}{2021}\natexlab{}.
\newblock \showarticletitle{Multimodal integration of human-like attention in
  visual question answering}.
\newblock \bibinfo{journal}{\emph{Computing Research Repository}}
  (\bibinfo{year}{2021}).
\newblock
\showeprint[arXiv]{2109.13139}


\bibitem[Sood et~al\mbox{.}(2020a)]%
        {sood2020interpreting}
\bibfield{author}{\bibinfo{person}{Ekta Sood}, \bibinfo{person}{Simon Tannert},
  \bibinfo{person}{Diego Frassinelli}, \bibinfo{person}{Andreas Bulling}, {and}
  \bibinfo{person}{Ngoc~Thang Vu}.} \bibinfo{year}{2020}\natexlab{a}.
\newblock \showarticletitle{Interpreting attention models with human visual
  attention in machine reading comprehension}. In
  \bibinfo{booktitle}{\emph{Proceedings of the 24th Conference on Computational
  Natural Language Learning}}. \bibinfo{address}{Online},
  \bibinfo{pages}{12--25}.
\newblock


\bibitem[Sood et~al\mbox{.}(2020b)]%
        {Sood2020ImprovingAttention}
\bibfield{author}{\bibinfo{person}{Ekta Sood}, \bibinfo{person}{Simon Tannert},
  \bibinfo{person}{Philipp M{\"u}ller}, {and} \bibinfo{person}{Andreas
  Bulling}.} \bibinfo{year}{2020}\natexlab{b}.
\newblock \showarticletitle{Improving natural language processing tasks with
  human gaze-guided neural attention}. In \bibinfo{booktitle}{\emph{Proceedings
  of the Conference on Neural Information Processing Systems}},
  Vol.~\bibinfo{volume}{33}. \bibinfo{address}{Online},
  \bibinfo{pages}{6327--6341}.
\newblock


\bibitem[Sutskever et~al\mbox{.}(2014)]%
        {sutskever2014sequence}
\bibfield{author}{\bibinfo{person}{Ilya Sutskever}, \bibinfo{person}{Oriol
  Vinyals}, {and} \bibinfo{person}{Quoc~V Le}.}
  \bibinfo{year}{2014}\natexlab{}.
\newblock \showarticletitle{Sequence to sequence learning with neural
  networks}. In \bibinfo{booktitle}{\emph{Proceedings of the Conference on
  Neural Information Processing Systems}}, Vol.~\bibinfo{volume}{27}.
\newblock


\bibitem[Swets et~al\mbox{.}(2008)]%
        {swets2008underspecification}
\bibfield{author}{\bibinfo{person}{Benjamin Swets}, \bibinfo{person}{Timothy
  Desmet}, \bibinfo{person}{Charles Clifton}, {and} \bibinfo{person}{Fernanda
  Ferreira}.} \bibinfo{year}{2008}\natexlab{}.
\newblock \showarticletitle{Underspecification of syntactic ambiguities:
  Evidence from self-paced reading}.
\newblock \bibinfo{journal}{\emph{Memory \& Cognition}}  \bibinfo{volume}{36}
  (\bibinfo{year}{2008}), \bibinfo{pages}{201--216}.
\newblock


\bibitem[Tabor and Hutchins(2004)]%
        {tabor2004evidence}
\bibfield{author}{\bibinfo{person}{Whitney Tabor} {and} \bibinfo{person}{Sean
  Hutchins}.} \bibinfo{year}{2004}\natexlab{}.
\newblock \showarticletitle{Evidence for self-organized sentence processing:
  digging-in effects.}
\newblock \bibinfo{journal}{\emph{Journal of Experimental Psychology: Learning,
  Memory, and Cognition}} \bibinfo{volume}{30}, \bibinfo{number}{2}
  (\bibinfo{year}{2004}), \bibinfo{pages}{431}.
\newblock


\bibitem[Takmaz et~al\mbox{.}(2020)]%
        {takmaz-etal-2020-generating}
\bibfield{author}{\bibinfo{person}{Ece Takmaz}, \bibinfo{person}{Sandro
  Pezzelle}, \bibinfo{person}{Lisa Beinborn}, {and} \bibinfo{person}{Raquel
  Fern{\'a}ndez}.} \bibinfo{year}{2020}\natexlab{}.
\newblock \showarticletitle{{G}enerating image descriptions via sequential
  cross-modal alignment guided by human gaze}. In
  \bibinfo{booktitle}{\emph{Proceedings of the 2020 Conference on Empirical
  Methods in Natural Language Processing}}. \bibinfo{address}{Online},
  \bibinfo{pages}{4664--4677}.
\newblock


\bibitem[Theis et~al\mbox{.}(2016)]%
        {theis2016a}
\bibfield{author}{\bibinfo{person}{Lucas Theis}, \bibinfo{person}{A{\"a}ron
  van~den Oord}, {and} \bibinfo{person}{Matthias Bethge}.}
  \bibinfo{year}{2016}\natexlab{}.
\newblock \showarticletitle{A note on the evaluation of generative models}. In
  \bibinfo{booktitle}{\emph{Proceedings of the 4th International Conference on
  Learning Representations, {ICLR}, San Juan, Puerto Rico}}.
\newblock


\bibitem[Wang et~al\mbox{.}(2019)]%
        {wang2019new}
\bibfield{author}{\bibinfo{person}{Xiaoming Wang}, \bibinfo{person}{Xinbo
  Zhao}, {and} \bibinfo{person}{Jinchang Ren}.}
  \bibinfo{year}{2019}\natexlab{}.
\newblock \showarticletitle{A new type of eye movement model based on recurrent
  neural networks for simulating the gaze behavior of human reading}.
\newblock \bibinfo{journal}{\emph{Complexity}}  \bibinfo{volume}{2019}
  (\bibinfo{year}{2019}), \bibinfo{numpages}{12}~pages.
\newblock


\end{thebibliography}

\appendix
\section{Appendix}
\subsection{Hyperparameter Tuning}
\label{sec:hyper-tune}
Table~\ref{tab: hp} shows the hyperparameter search grid used to find an optimal model configuration. The last column indicates the best found values.
\begin{table*}[h!]
\caption{Parameter used for hyperparameter optimization.}
\label{tab: hp}
\begin{center}
    \begin{tabular}{l|l|l}
    \toprule
    Parameter       & Search space  & Best value           \\\hline
    number of LSTM / BiLSTM layers   & \{2, 4, 8, 12\} & 8 / 8\\
    LSTM / BiLSTM units       & \{16, 32, 64, 128, 256, 512\} & 128 / 64\\
    number of dense layers & \{1, 2, 4, 8\} & 4\\
    number of dense units & \{64, 128, 256, 512, 1024\} & {512, 256, 256, 256}\\
    embedding dropout rate  & \{0, 0.1, 0.2, 0.3, 0.4, 0.5\} &0.4\\
    LSTM / BiLSTM dropout rate & \{0, 0.1, 0.2, 0.3, 0.4, 0.5\} &0.2\\
    dense dropout rate & \{0, 0.1, 0.2, 0.3, 0.4, 0.5\} &0.2 \\
    attention window size    & \{1, 2, 3, 4, 5\} & 1\\
    \bottomrule
    \end{tabular}
\end{center}
\end{table*}

\subsection{Scanpath Similarity Scores}
\label{sec:multimatch}
Here we present results based on the scanpath similarity metric MultiMatch~\citep{jarodzka2010vector} for the within-data set evaluation. We adhere to the same evaluation protocol as outlined in the main article, as described in Section~\ref{sec:within-data-eval}. The MultiMatch metric is designed to measure the similarity of scanpaths represented as 2D geometric vectors. Since our work is on comparing scanpaths generated while reading text, we represent each saccade vector in a 1D space using the word index as the start and end positions. In Table~\ref{tab: multimatch_score}, we report three measures included in the MultiMatch metric: shape, length, and position. These measures are calculated using the publicly available python package \footnote{\url{https://github.com/adswa/multimatch_gaze}}. We choose not to include these measures in the main article due to concerns about their validity raised in a previous study~\cite{kummerer2021state}. This study demonstrated that incorrect models can systematically score higher than the ground truth model. Moreover, the adaptation of the MultiMatch metric to a one-dimensional space, which is necessary for tasks such as the comparison of scanpaths generated during text reading, remains an area that requires further investigation.
\begin{table*}[]
\caption{MultiMatch scores on the BSC and CELER L1 datasets for 5-fold CV/random resampling for the New Reader split, New Sentence split, and New Reader / New Sentence split, respectively. We follow the same evaluation procedure as for the NLD metric in the main article: for each ground truth human scanpath, we compute its similarity to a sampled scanpath from the model. }

\label{tab: multimatch_score}
\small
\begin{center}
    \begin{tabular}{l|l|l|l|l|l}
    \toprule
             &                  &              & \multicolumn{3}{c}{MultiMatch}\\
    Data set & Train/test split &  Model       & Shape $\uparrow$ &Length $\uparrow$  &Position $\uparrow$        \\\hline
    BSC   & New Sentence & Uniform               &0.9423 $\pm$ 0.06 & 0.9021 $\pm$ 0.102 & 0.684 $\pm$ 0.12 \\
(Chinese) &              &Train-label-dist      &0.988 $\pm$ 0.013  &0.9794 $\pm$ 0.025 & 0.8021 $\pm$ 0.118 \\
          &              & E-Z Reader           & 0.968 $\pm$ 0.017 & 0.9566 $\pm$ 0.025 & 0.7795 $\pm$ 0.13 \\
          &              &\citet{nilsson2009learning}   & 0.9916 $\pm$ 0.012   &0.986 $\pm$ 0.022  & 0.8352 $\pm$ 0.105    \\
          &              &\citet{nilsson2011entropy}   & 0.993 $\pm$ 0.011  & 0.9887 $\pm$ 0.02  & \textbf{0.8473 $\pm$ 0.097}\\
          &              &Eyettention (Ours)        &\textbf{0.9934 $\pm$ 0.011}   & \textbf{0.9893 $\pm$ 0.02}  & 0.8427 $\pm$ 0.095\\ \cline{3-6}
          &              &\textit{Human} & 0.9932 $\pm$ 0.011 & 0.9894 $\pm$ 0.019 & 0.8625 $\pm$ 0.09 \\
          \cline{2-6}

          & New Reader   & Uniform               &0.9418 $\pm$ 0.06  & 0.9011 $\pm$ 0.104 &0.6862 $\pm$ 0.119 \\
          &             &Train-label-dist      &0.9879 $\pm$ 0.0130   &0.9792 $\pm$ 0.025  &0.8031 $\pm$ 0.117\\
          &              & E-Z Reader &0.968 $\pm$ 0.017 & 0.9564 $\pm$ 0.026 & 0.7785 $\pm$ 0.131 \\
          &              &\citet{nilsson2009learning}   & 0.9912 $\pm$ 0.012  &0.9854 $\pm$ 0.022 &0.834 $\pm$ 0.107    \\
          &              &\citet{nilsson2011entropy}   & 0.9929 $\pm$ 0.011  &0.9886 $\pm$ 0.02 &\textbf{0.8482 $\pm$ 0.098}\\
          &              &Eyettention (Ours)        &\textbf{0.9937 $\pm$ 0.011}   & \textbf {0.9896$\pm$ 0.019} & 0.8465 $\pm$ 0.094 \\ \cline{3-6}
          &              &\textit{Human} &0.9932 $\pm$ 0.011 &0.9892 $\pm$ 0.02 &0.8592 $\pm$ 0.09\\
          \cline{2-6}

          & New Reader/   & Uniform               &0.9476 $\pm$ 0.054  & 0.9121 $\pm$ 0.09 & 0.6752$\pm$ 0.115\\
          & New Sentence  &Train-label-dist      &0.9888 $\pm$ 0.013   &0.9811 $\pm$ 0.024 &0.7951 $\pm$ 0.12\\
          &   & E-Z Reader &0.9682 $\pm$ 0.017 & 0.9575 $\pm$ 0.024 & 0.7842 $\pm$ 0.132 \\
          &              &\citet{nilsson2009learning}   & 0.992 $\pm$ 0.012   & 0.9865 $\pm$ 0.022 &0.8342 $\pm$ 0.104    \\
          &              &\citet{nilsson2011entropy}   & \textbf{0.9941 $\pm$ 0.01}  & \textbf{0.9908 $\pm$ 0.018} & 0.8443 $\pm$ 0.096\\
          &              &Eyettention (Ours)        &0.9939 $\pm$ 0.011   & 0.9901 $\pm$ 0.019 & \textbf{0.8453 $\pm$ 0.091}\\ \cline{3-6}
          &              &\textit{Human} &0.9945 $\pm$ 0.01&0.9916 $\pm$ 0.018 & 0.86 $\pm$ 0.089\\\hline

CELER~L1   & New Sentence & Uniform               &0.9345 $\pm$ 0.067 & 0.8873 $\pm$ 0.122  &0.6129 $\pm$ 0.139\\
(English) &              &Train-label-dist      &0.9636 $\pm$ 0.032   &0.9431 $\pm$ 0.058 &0.7452 $\pm$ 0.143 \\
          &              & E-Z Reader &0.9680 $\pm$ 0.0265& 0.942 $\pm$ 0.020 &0.752 $\pm$ 0.235\\
          &              & SWIFT&0.9674 $\pm$ 0.032&0.9498 $\pm$ 0.045&0.7822 $\pm$ 0.062\\
          &              &\citet{nilsson2009learning}   & 0.9678 $\pm$ 0.031   &0.9513 $\pm$ 0.056 & 0.7751 $\pm$ 0.135    \\
          &              &\citet{nilsson2011entropy}   & 0.9679 $\pm$ 0.003  &0.9513 $\pm$ 0.056 & 0.7907  $\pm$ 0.128\\
          &              &Eyettention (Ours)        &\textbf{0.9689 $\pm$ 0.03}   & \textbf {0.9544$\pm$ 0.054} & \textbf{0.809 $\pm$ 0.109} \\ \cline{3-6}
          &              &\textit{Human} &0.9717 $\pm$ 0.029&\textit{0.9577 $\pm$ 0.053} &0.8348 $\pm$ 0.106\\
          \cline{2-6}

          & New Reader  & Uniform               &0.9366 $\pm$ 0.065 & 0.8916 $\pm$ 0.119 &0.6139 $\pm$ 0.138 \\
          &              &Train-label-dist      &0.9639 $\pm$ 0.031   &0.9437 $\pm$ 0.057 &0.7434 $\pm$ 0.143 \\
          &              & E-Z Reader &0.9613 $\pm$ 0.026 & 0.9496 $\pm$ 0.017 & 0.7543 $\pm$ 0.2201\\
          &              & SWIFT& 0.9675 $\pm$ 0.030 & 0.9390 $\pm$ 0.062 & 0.7309 $\pm$ 0.225\\
          &              &\citet{nilsson2009learning}   & 0.9676 $\pm$ 0.031  &0.9502 $\pm$ 0.057 &0.7779 $\pm$ 0.132     \\
          &              &\citet{nilsson2011entropy}   & 0.9684 $\pm$ 0.030  &0.9518 $\pm$ 0.055 &0.7909 $\pm$ 0.128\\
          &              &Eyettention (Ours)        &\textbf{0.9695 $\pm$ 0.03}   & \textbf {0.9548$\pm$ 0.054}  &\textbf{0.8098 $\pm$ 0.11}  \\ \cline{3-6}
          &              &\textit{Human} &0.9713 $\pm$ 0.03&\textit{0.9567 $\pm$ 0.054} &0.8367 $\pm$ 0.106\\
          \cline{2-6}

          & New Reader/   & Uniform               &0.9355 $\pm$ 0.065 & 0.8879 $\pm$ 0.121 & 0.616 $\pm$ 0.14\\
          & New Sentence   &Train-label-dist      &0.9649 $\pm$ 0.031  &0.9446 $\pm$ 0.058 &0.7492 $\pm$ 0.139 \\
          &   & E-Z Reader &0.9603 $\pm$ 0.026 & 0.9506 $\pm$ 0.021 & 0.7523 $\pm$ 0.203\\
          &              & SWIFT&0.9590 $\pm$ 0.030&0.9330 $\pm$ 0.059& 0.7301 $\pm$ 0.230\\
          &              &\citet{nilsson2009learning}   & 0.9696 $\pm$ 0.029   &\textbf{0.9552 $\pm$ 0.053} & 0.7801 $\pm$ 0.136   \\
          &              &\citet{nilsson2011entropy}   & \textbf{0.9699 $\pm$ 0.029}  &0.9542 $\pm$ 0.053  & 0.7912  $\pm$ 0.127\\
          &              &Eyettention (Ours)        &0.9689 $\pm$ 0.029   & 0.9537$\pm$ 0.053 & \textbf{0.8057 $\pm$ 0.108}\\ \cline{3-6}
          &              &\textit{Human} & 0.9737 $\pm$ 0.028 &0.9611 $\pm$ 0.051 &\textit{0.8389 $\pm$ 0.104}\\
    
    \bottomrule
    \end{tabular}
\end{center}
\end{table*}

\subsection{Additional Ablation Studies: Effect of Asymmetric Attention Window}
\label{sec:asym-window}
We present the results of Eyettention model using an asymmetric attention window to account for the asymmetry of the perceptual span while reading. We first experimented with a right-skewed window, which spans more words to the right, and applied a Gaussian kernel with varying standard deviations for
the left and right parts of the windows, where the right window had a larger standard deviation, as designed in the
SWIFT model~\citep{engbert2005swift}. In addition, we also experimented with shifting the center of the
Gaussian kernel one word to the right of the current fixation position. We incorporated different parameter
combinations and their corresponding results into Table~\ref{tab: asym-ablation}. With the exception of the New Reader Split for the BSC dataset, we did not observe any significant improvements over our original model. For the English dataset, many of the parameter combinations resulted in significantly worse performance compared to the original model.

\begin{table*}[]
\caption{Experimental results of Eyettention model with a symmetric window compared to the model using an asymmetric window. We report NLL $\pm$ standard error for 5-fold CV/random resampling for the New Reader split, New Sentence split, and New Reader / New Sentence split, respectively. Lower scores indicate better performance. The dagger $\dagger$ indicates models significantly worse than the original Eyettention model with the symmetric window and the asterisk * indicates models significantly better than the original Eyettention model. D denotes window width, $\mu$ and $\sigma$ represent the mean and standard deviation of the Gaussian kernel, respectively. $f_i$ indicates the word index of the current fixation. To accommodate space limitations, we use the abbreviation \emph{G.} to refer to \emph{Gaussian}, and \emph{D$_l$} and \emph{D$_r$} to represent the left and right window widths, respectively.}
\label{tab: asym-ablation}
\small
\begin{center}
    \begin{tabular}{l|l|l|l|l}
    \toprule
    && \multicolumn{3}{c}{Train/test split (NLL$\downarrow$)}\\
    Data Set   & Model                              & New Sentence                  & New Reader                 & New Reader / \\
               &                                    &                               &                            &New Sentence          \\\hline
        BSC   & Eyettention (original)     &1.856 $\pm$ 0.007              & 1.875 $\pm$ 0.007          & 1.840 $\pm$ 0.017 \\
        \cline{2-5}
 (Chinese)    &Eyettention w/ right skewed G. window &&&\\
   &D$_{l}$=1, D$_{r}$=2, $\mu$=$f_i$, $\sigma$=D/2        & 1.851 $\pm$ 0.007             & 1.874 $\pm$ 0.007          & 1.845 $\pm$ 0.016\\
              &D$_{l}$=1, D$_{r}$=2, $\mu$=$f_i$, $\sigma$=D       & 1.848 $\pm$ 0.007             & 1.851$\pm$ 0.007*          & 1.835 $\pm$ 0.016\\
              &D$_{l}$=1, D$_{r}$=3, $\mu$=$f_i$, $\sigma$=D/2        & 1.845 $\pm$ 0.007             & 1.862 $\pm$ 0.007          & 1.843 $\pm$ 0.017\\
              &D$_{l}$=1, D$_{r}$=3, $\mu$=$f_i$, $\sigma$=D       & 1.849 $\pm$ 0.007             & 1.856 $\pm$ 0.007          & 1.84 $\pm$ 0.017\\
              &D$_{l}$=1, D$_{r}$=3, $\mu$=$f_i$, $\sigma$=2D        & 1.848 $\pm$ 0.007             & 1.846 $\pm$ 0.007*          & 1.859 $\pm$ 0.017\\
              \cline{2-5}

              &Eyettention w/ right shifted G. kernel    &             &          & \\
               &D$_{l}$=1, D$_{r}$=1, $\mu$=$f_i$+1, $\sigma$=D$_{l}$        & 1.845 $\pm$ 0.007             & \textbf{1.843 $\pm$ 0.007}*          & \textbf{1.831 $\pm$ 0.017}\\
               &D$_{l}$=1, D$_{r}$=1, $\mu$=$f_i$+1, $\sigma$=2D$_{l}$        & \textbf{1.843 $\pm$ 0.007}             & 1.853 $\pm$ 0.007*          & 1.833 $\pm$ 0.017\\
               &D$_{l}$=1, D$_{r}$=2, $\mu$=$f_i$+1, $\sigma$=D$_{l}$        & 1.847 $\pm$ 0.007             & 1.858 $\pm$ 0.007          & \textbf{1.831 $\pm$ 0.017}\\
               &D$_{l}$=1, D$_{r}$=2, $\mu$=$f_i$+1, $\sigma$=2D$_{l}$        & 1.844 $\pm$ 0.007             & 1.844 $\pm$ 0.007*          & 1.844 $\pm$ 0.017\\\hline\hline

    CELER~L1   &Eyettention (original)   &2.277 $\pm$ 0.005       & \textbf{2.267 $\pm$ 0.005}         &2.297 $\pm$ 0.011\\
    \cline{2-5}
    (English)  &Eyettention w/ right skewed G. window        &           &        & \\
      &D$_{l}$=1, D$_{r}$=2, $\mu$=$f_i$, $\sigma$=D/2         &  \textbf{2.272 $\pm$ 0.005}             & 2.276 $\pm$ 0.006          & 2.31 $\pm$ 0.011\\
                &D$_{l}$=1, D$_{r}$=2, $\mu$=$f_i$, $\sigma$=D)       & 2.322 $\pm$ 0.005$\dagger$             &  2.32 $\pm$ 0.005$\dagger$         & 2.338 $\pm$ 0.011$\dagger$\\
                &D$_{l}$=1, D$_{r}$=3, $\mu$=$f_i$, $\sigma$=D/2        & 2.311$\pm$ 0.005$\dagger$            & 2.307 $\pm$ 0.005$\dagger$          & 2.336$\pm$ 0.011$\dagger$\\
                &D$_{l}$=1, D$_{r}$=3, $\mu$=$f_i$, $\sigma$=D        & 2.326 $\pm$ 0.005$\dagger$             & 2.329 $\pm$ 0.005$\dagger$          & 2.347 $\pm$ 0.011$\dagger$\\
              &D$_{l}$=1, D$_{r}$=3, $\mu$=$f_i$, $\sigma$=2D        & 2.33 $\pm$ 0.005$\dagger$             & 2.321 $\pm$ 0.005$\dagger$          & 2.346 $\pm$ 0.011$\dagger$\\
                \cline{2-5}
                
              &Eyettention w/ right shifted G. kernel        &          &          & \\
               &D$_{l}$=1, D$_{r}$=1, $\mu$=$f_i$+1, $\sigma$=D$_{l}$        & 2.288 $\pm$ 0.005             & 2.282 $\pm$ 0.007          & 2.311 $\pm$ 0.011\\
               &D$_{l}$=1, D$_{r}$=1, $\mu$=$f_i$+1, $\sigma$=2D$_{l}$       & 2.289 $\pm$ 0.005             & 2.283 $\pm$ 0.005$\dagger$          & \textbf{2.294 $\pm$ 0.011}\\
               &D$_{l}$=1, D$_{r}$=2, $\mu$=$f_i$+1, $\sigma$=D$_{l}$        & 2.294 $\pm$ 0.005$\dagger$             & 2.298 $\pm$ 0.005$\dagger$          & 2.329 $\pm$ 0.011$\dagger$\\
               &D$_{l}$=1, D$_{r}$=2, $\mu$=$f_i$+1, $\sigma$=2D$_{l}$        & 2.314 $\pm$ 0.005$\dagger$             & 2.316 $\pm$ 0.005$\dagger$          & 2.336 $\pm$ 0.011$\dagger$\\
    \bottomrule 
    \end{tabular}

\end{center}
\end{table*}

\subsection{Performance wrt Saccade Length}
\label{sec:sac-length}
\begin{figure}[!ht]
    \subfloat[BSC dataset, New Sentence Split.\label{subfig:sac_type_BSC_NS}]{%
      \includegraphics[width=0.49\textwidth,keepaspectratio]{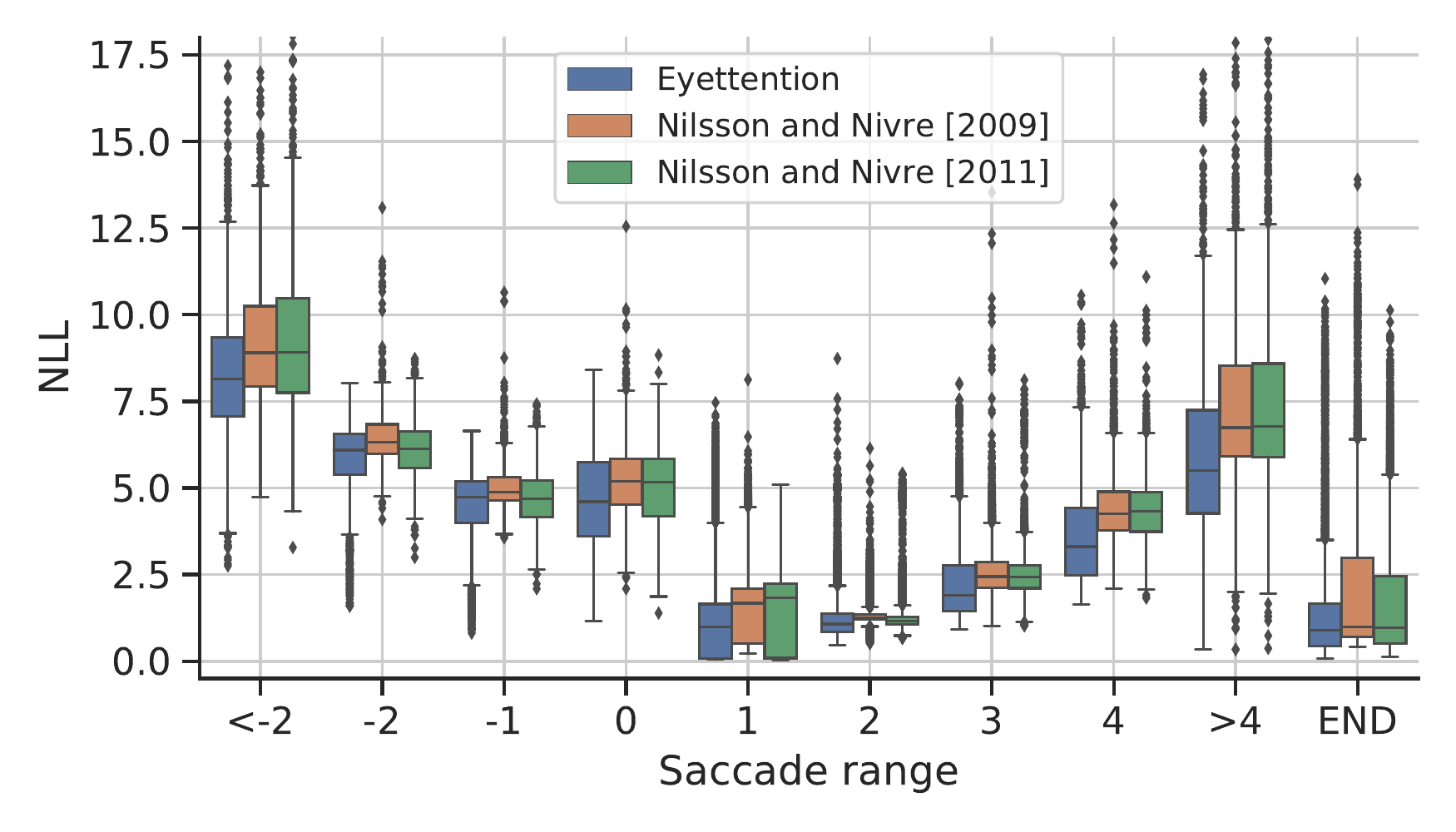}
    }
    \hfill
    \subfloat[CELER dataset, New Sentence Split.\label{subfig:sac_type_celer_NS}]{%
      \includegraphics[width=0.49\textwidth,keepaspectratio]{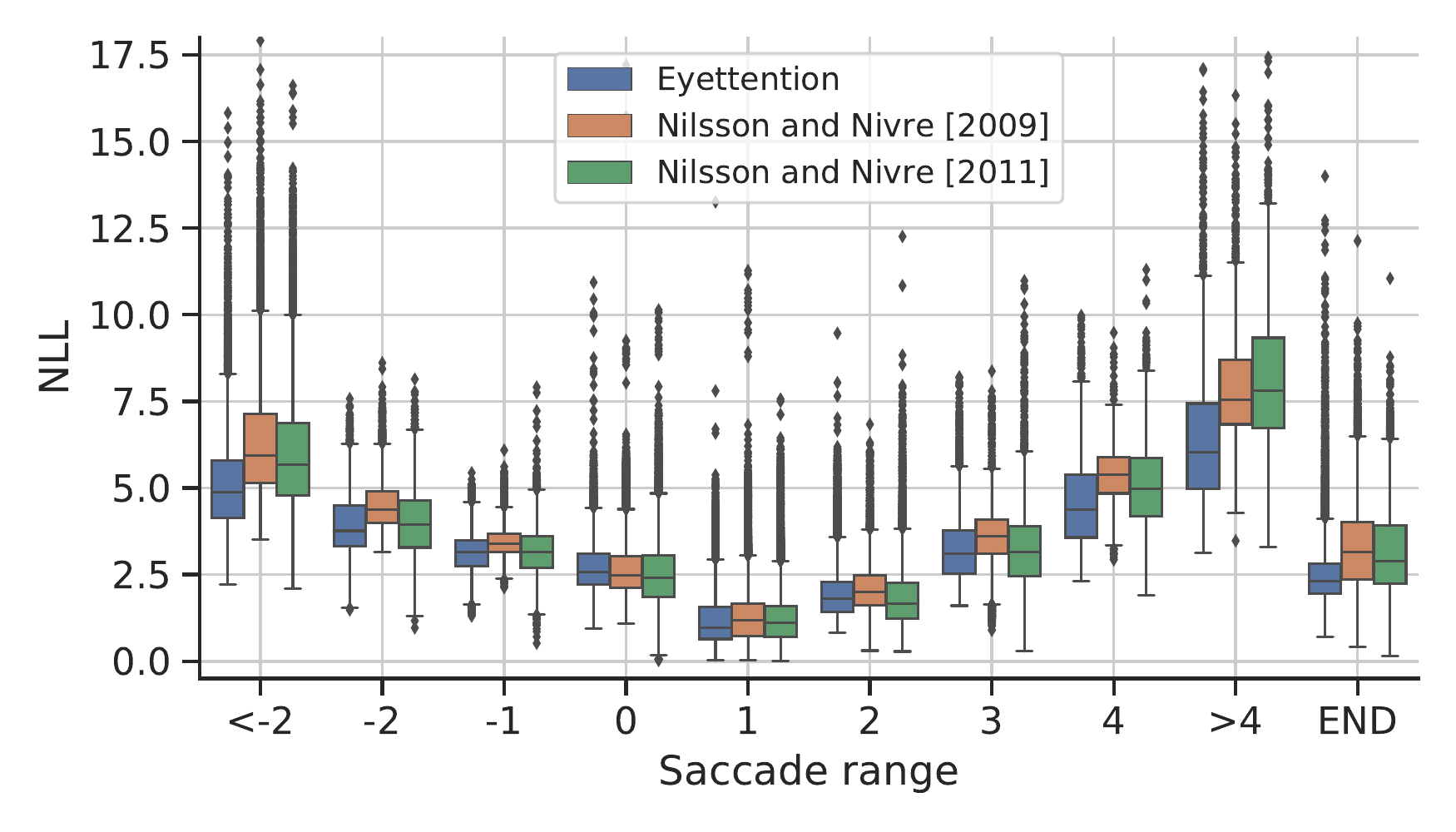}
    }
    \vfill
    \subfloat[BSC dataset, New Reader Split.\label{subfig:sac_type_BSC_NR}]{%
      \includegraphics[width=0.49\textwidth,keepaspectratio]{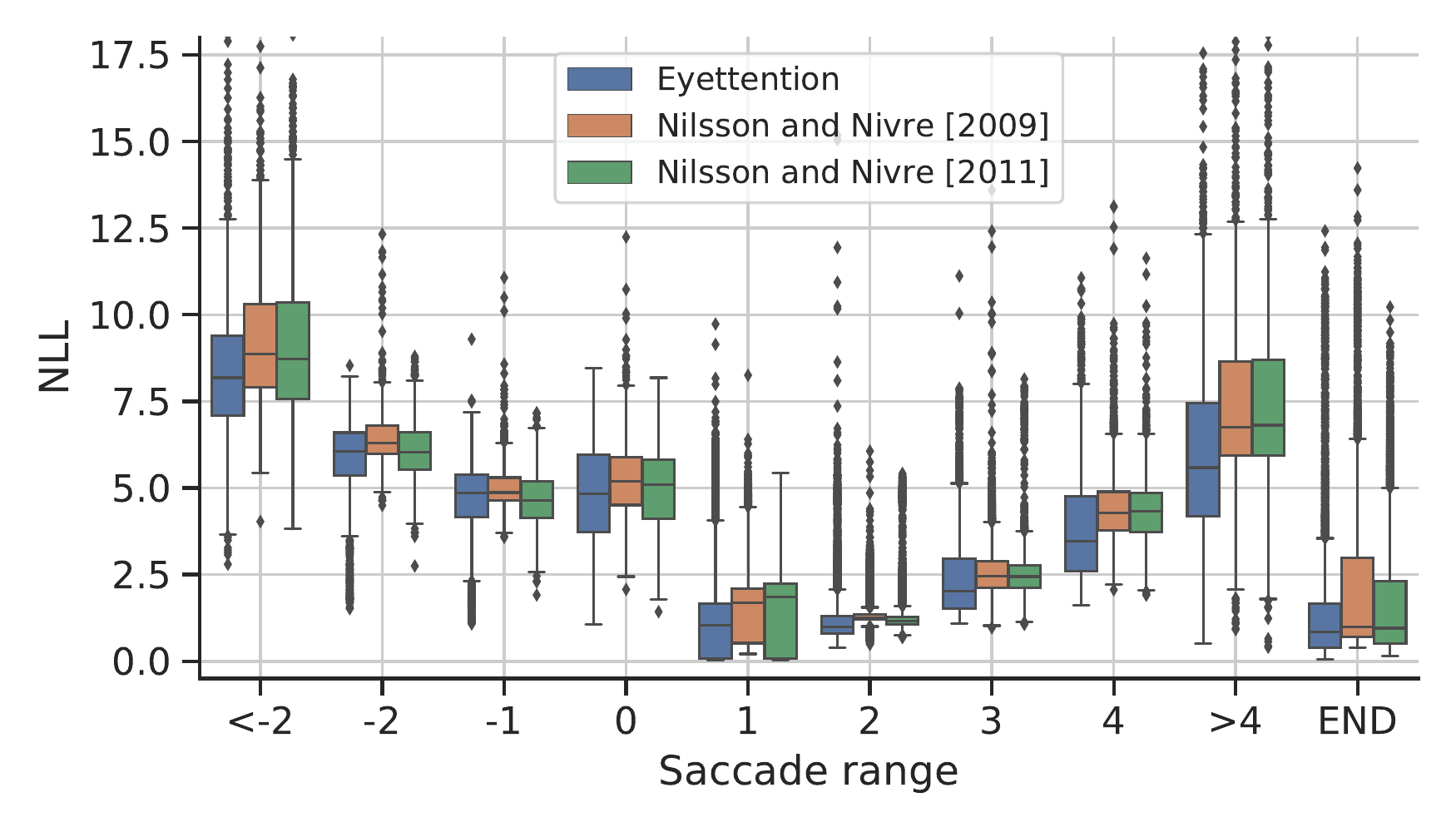}
    }
    \hfill
    \subfloat[CELER L1 dataset, New Reader Split.\label{subfig:sac_type_celer_NR}]{%
      \includegraphics[width=0.49\textwidth,keepaspectratio]{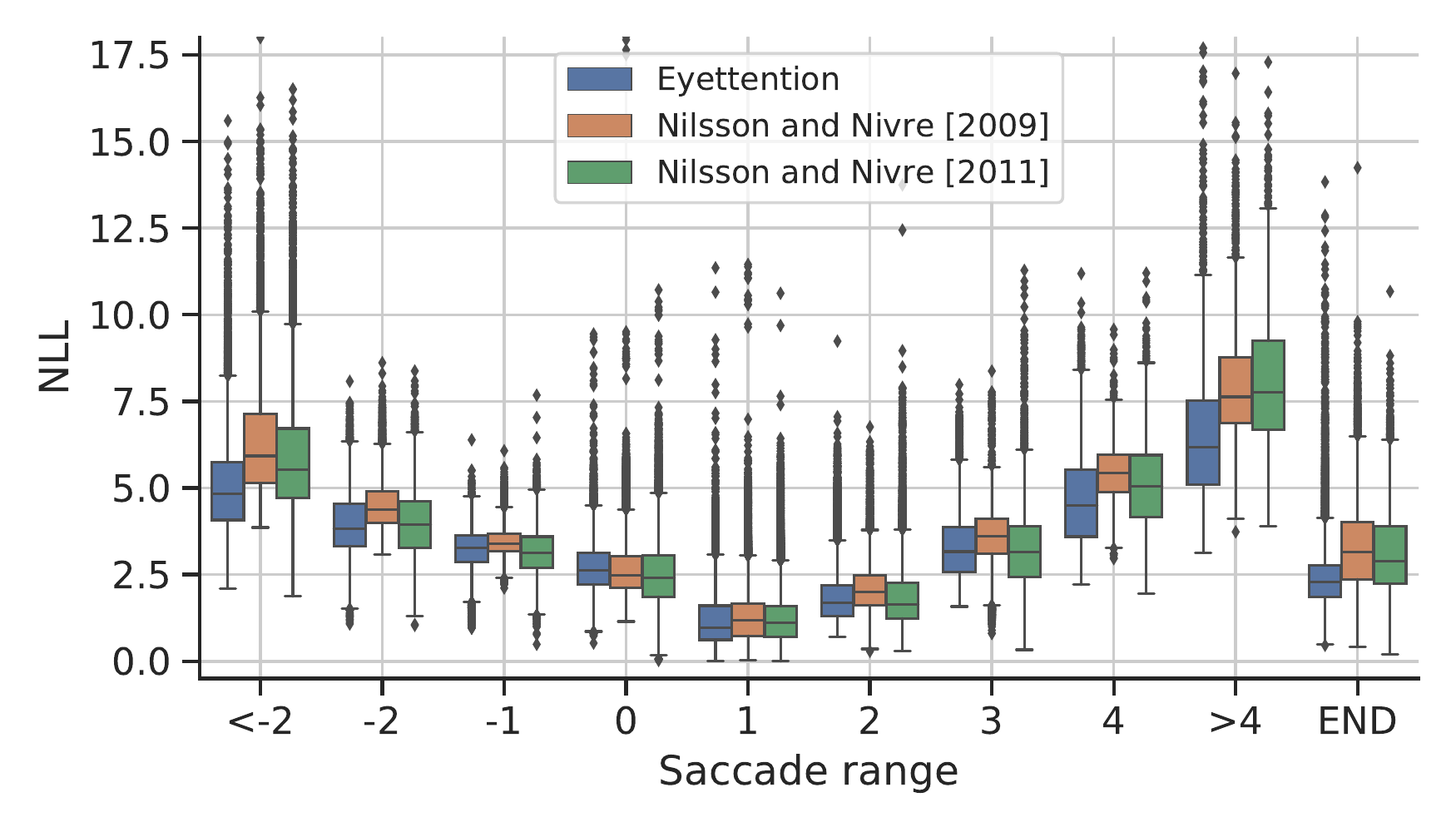}
    }
    \caption{NLL over different saccade ranges for the different methods on BSC and CELER L1 dataset.}
    \label{fig: sac_type_rest}
  \end{figure}

Figure~\ref{fig: sac_type_rest} shows the performance  of the investigated models as a function of the range of the next saccade for the New Sentence and New Reader split.

\subsection{Model Predictions Visualization}
\label{sec:pred-vis}
\begin{figure}[!t]
\begin{subfigure}[t]{1\textwidth}
\centering
\includegraphics[width=1\textwidth,keepaspectratio]{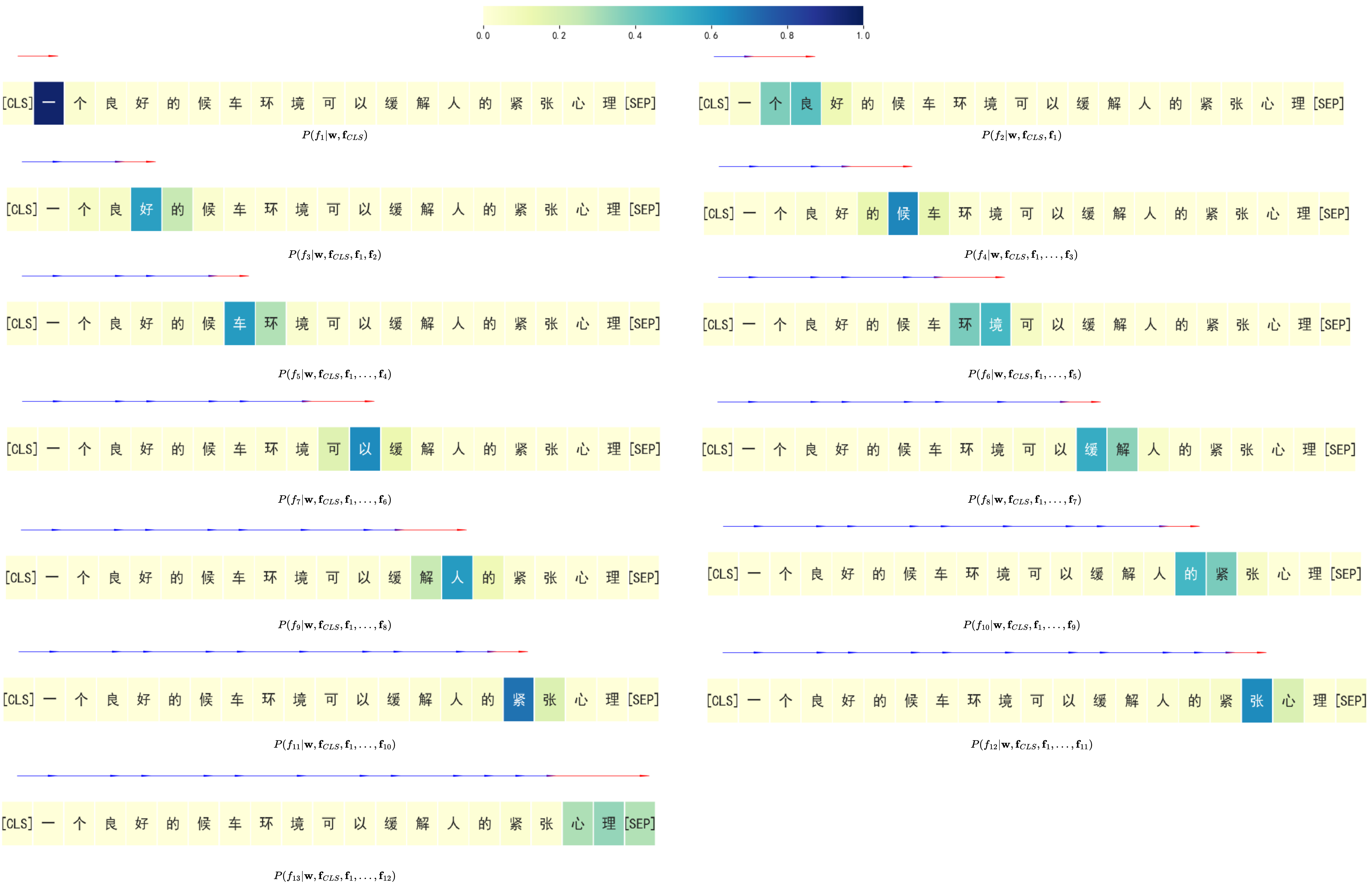}
\subcaption{BSC dataset}
\end{subfigure}
\begin{subfigure}[t]{0.9\textwidth}
\centering
\includegraphics[width=.99\textwidth,keepaspectratio]{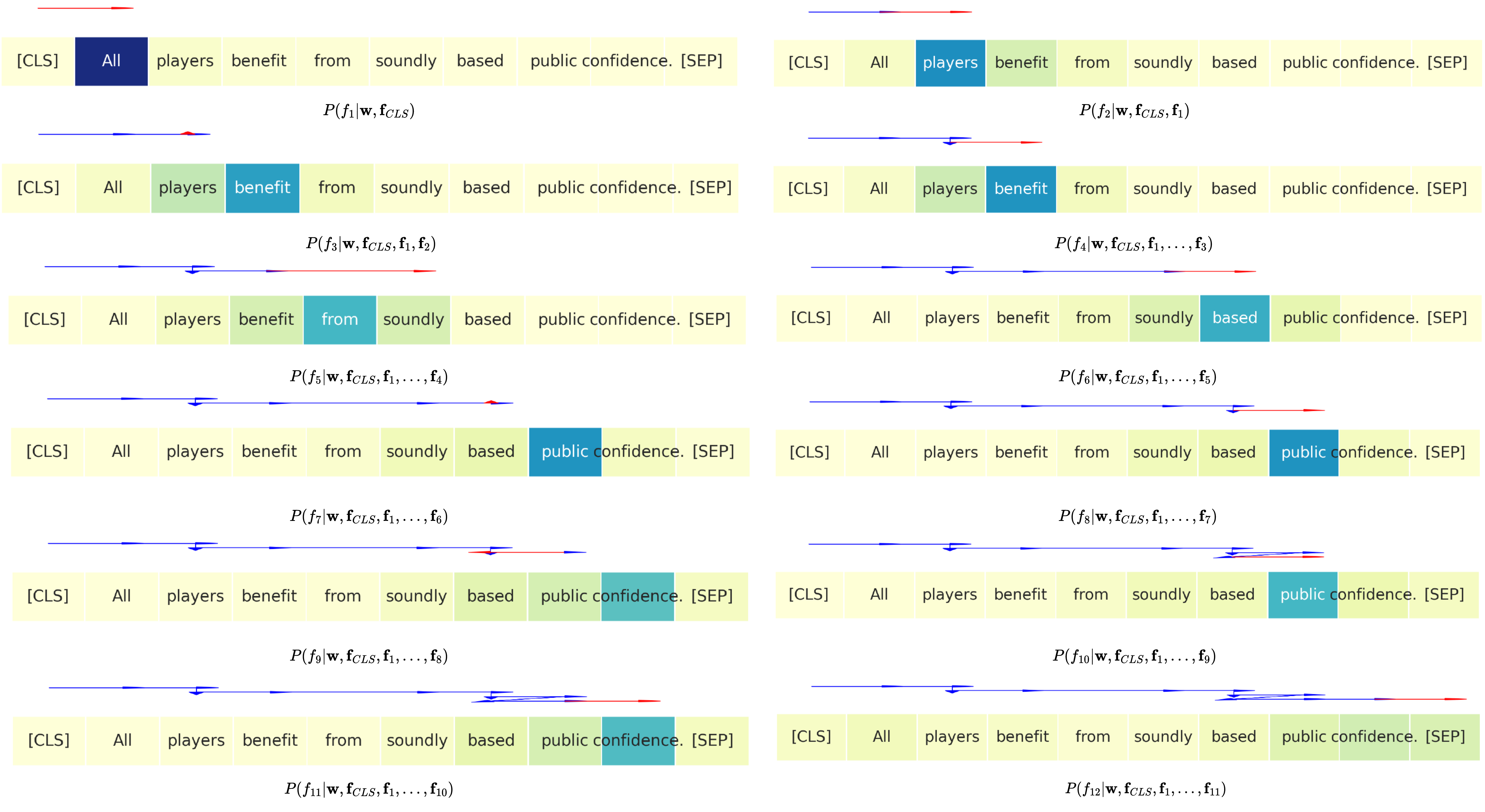}
\subcaption{CELER L1 dataset}
\end{subfigure}
\caption{The likelihood $P(f_i|\mathbf{W},\mathbf{f}_0,\dots,\mathbf{f}_{i-1})$ is displayed as a heatmap. The red arrows indicate the ground truth of the current step and the blue arrows indicate the previous fixation positions.}
\label{fig:pred_vis}
\end{figure}

Figure~\ref{fig:pred_vis} visualizes the predictions of the Eyettention model for example scanpaths.
\end{document}